\documentclass[lettersize,journal]{IEEEtran}

\usepackage[caption=false,font=normalsize,labelfont=sf,textfont=sf]{subfig}
\usepackage{textcomp}
\usepackage[dvipsnames]{xcolor}

\usepackage[utf8]{inputenc} % allow utf-8 input
\usepackage[T1]{fontenc}    % use 8-bit T1 fonts
\usepackage{url}            % simple URL typesetting
\usepackage{booktabs}       % professional-quality tables
\usepackage{amsfonts}       % blackboard math symbols
\usepackage{nicefrac}       % compact symbols for 1/2, etc.
\usepackage{microtype}      % microtypography

\usepackage{graphicx}
\usepackage{amssymb}% http://ctan.org/pkg/amssymb
\usepackage{pifont}% http://ctan.org/pkg/pifont
\usepackage{multirow}
\usepackage{amsmath}
\usepackage{color, colortbl}
\usepackage{caption}

\usepackage{siunitx}
\definecolor{citecolor}{HTML}{0071bc}

\definecolor{Gray}{gray}{0.9}

\newcommand{\KGnote}[1]{{\color{red}{\bf KG: }#1}}

\newcommand{\CAnote}[1]{{\color{brown}{\bf CA: }#1}}

\newcommand{\cmark}{\ding{51}}%
\newcommand{\xmark}{\ding{55}}%

\newcommand{\RNum}[1]{\uppercase\expandafter{\romannumeral #1\relax}}

\renewcommand{\CAnote}[1]{}
\renewcommand{\KGnote}[1]{}

\usepackage{amsmath,amsfonts}
\usepackage[algo2e]{algorithm2e} 
\usepackage{algorithmic}
\usepackage{algorithm}
\usepackage{array}
\usepackage[caption=false,font=normalsize,labelfont=sf,textfont=sf]{subfig}
\usepackage{textcomp}
\usepackage{stfloats}
\usepackage{url}
\usepackage{verbatim}
\usepackage{graphicx}
\usepackage{cite}

\def\BibTeX{{\rm B\kern-.05em{\sc i\kern-.025em b}\kern-.08em
    T\kern-.1667em\lower.7ex\hbox{E}\kern-.125emX}}
\usepackage{balance}
\usepackage{times}
\usepackage{fancyhdr,graphicx,amsmath,amssymb}

\usepackage{multirow} 
\usepackage{arydshln}

\usepackage{url}
\usepackage{graphicx}
\usepackage{booktabs} 

\usepackage{xcolor}
\usepackage{hyperref}
\definecolor{deepgreen}{rgb}{0.0, 0.5, 0.0}

\begin{document}
\include{pythonlisting}

\title{SOWing Information: Cultivating Contextual Coherence with MLLMs in Image Generation}

\author{Yuhan Pei, Ruoyu Wang, Yongqi Yang, Ye Zhu, Olga Russakovsky, Yu Wu

\thanks{Y. Pei, R. Wang, Y. Yang, and Y. Wu are with the School of Computer Science, Wuhan University, China. E-mail: {yuhanpei, wangruoyu, yongqiyang, wuyucs}@whu.edu.cn}
\thanks{Y. Zhu, O. Russakovsky are with the Department of Computer Science, Princeton University, Princeton, NJ 08540 USA. E-mail: {yezhu, olgarus}@princeton.edu.}
}
\maketitle

\begin{abstract}
 
Originating from the diffusion phenomenon in physics, which describes the random movement and collisions of particles, diffusion generative models simulate a random walk in the data space along the denoising trajectory. 
This allows information to diffuse across regions, yielding harmonious outcomes.
However, the chaotic and disordered nature of information diffusion in diffusion models often results in undesired interference between image regions, causing degraded detail preservation and contextual inconsistency.
In this work, we address these challenges by reframing disordered diffusion as a powerful tool for text-vision-to-image generation (TV2I) tasks, achieving pixel-level condition fidelity while maintaining visual and semantic coherence throughout the image.
We first introduce Cyclic One-Way Diffusion (COW), which provides an efficient unidirectional diffusion framework for precise information transfer while minimizing disruptive interference. 
Building on COW, we further propose Selective One-Way Diffusion (SOW), which utilizes Multimodal Large Language Models (MLLMs) to clarify the semantic and spatial relationships within the image. Based on these insights, SOW combines attention mechanisms to dynamically regulate the direction and intensity of diffusion according to contextual relationships.
Extensive experiments demonstrate the untapped potential of controlled information diffusion, offering a path to more adaptive and versatile generative models in a learning-free manner.
Project page: \href{https://pyh-129.github.io/SOW/}{\texttt{https://pyh-129.github.io/SOW/}}
\end{abstract}


\section{Introduction}\label{sec:intro}

\IEEEPARstart{I}n physics, the diffusion phenomenon describes the movement of particles from an area of higher concentration to a lower concentration area till an equilibrium is reached~\cite{philibert2006one}. It represents a stochastic random walk of molecules to explore the space, from which originates the diffusion generative models~\cite{nonequilibrium}. Similar to the physical diffusion process, it is widely acknowledged that the diffusion generative model in machine learning also stimulates a random walk in the data space \cite{song2020score,mlt_sde,lu2022dpm}, however, it is less obvious how the diffusion models stimulate the information diffusion for real-world data along its walking trajectory. In this work, we start by investigating the diffusion phenomenon in diffusion models for image synthesis, namely \emph{``diffusion in diffusion''}, during which the pixels within a single image exchange and interact with each other, ultimately achieving a harmonious state in the data space (see Sec.~\ref{subsec:diffusion_in_diffusion}).

\begin{figure*}[t!]
    \centering
    \includegraphics[width=0.9\textwidth]{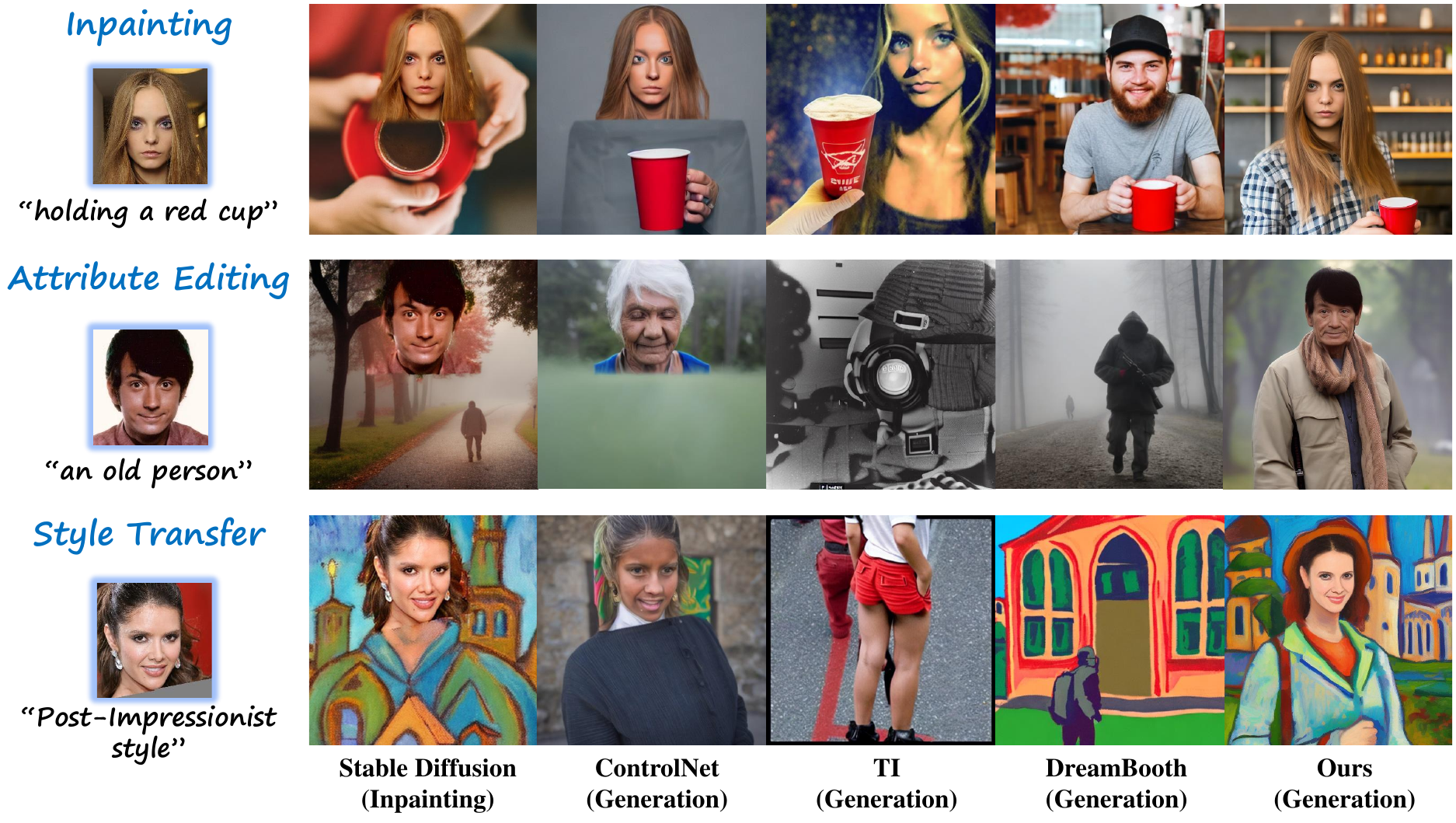}
    \caption{Comparisons with existing methods~\cite{rombach2022ldm,controlnet,gal2022ti,ruiz2022dreambooth} for maintaining the fidelity of text and visual conditions in different application scenarios. We consistently achieve superior fidelity to both text and visual conditions in all three settings. In contrast, other learning-based approaches struggle to attain the same level of performance in diverse scenarios. 
    }
    \label{fig:teaser}
    \vspace{-0.07in}
\end{figure*}

Despite their success, diffusion models face inherent limitations that can hinder their effectiveness in practical applications. One significant challenge is unwanted interference. For instance, tasks like image inpainting can be viewed as strictly unidirectional information diffusion, where information propagates from a known region to an unknown region while preserving the pixel-level integrity of the known content. However, uncontrolled diffusion in this scenario can lead to the intrusion of information from unknown regions into known regions, potentially disrupting the known content and introducing artifacts in the inpainted image.
Additionally, in physical diffusion, particles move randomly without specific destinations. Similarly, generative models propagate information across images without fully understanding the roles and relationships of different regions. 
This indiscriminate information transfer can result in incorrect information leakage and insufficient guidance, causing suboptimal information distribution. Consequently, current generative models may produce outputs that are visually fragmented and semantically incoherent (see Fig.~\ref{fig:teaser}). 
Furthermore, they may also encounter issues such as catastrophic neglect, attribute misalignment, and attribute leakage\cite{chefer2023attend,Li2023divide,huang2023t2icompbench}. 
This underscores the critical demand for more sophisticated generative models that can \emph{more adeptly grasp and regulate the intricate dynamics between different components in images}.

In this work, we focus on a multimodal generation task setting to synthesize images that successfully incorporate both semantic-Text and pixel-Vision conditioning (TV2I), allowing users to create more customized images.
Most existing methods~\cite{gal2022ti, ruiz2022dreambooth, dong2022dreamartist, controlnet} condition on text-vision input by brute-force learning, where they incorporate the visual condition into pre-trained T2I models by introducing an additional finetuning process to minimize the reconstruction error. 
Despite their abilities to capture the high-level semantics of extra visual conditioning, these methods often struggle with retaining low-level visual details in pixels as illustrated in Fig.~\ref{fig:teaser}.
Additionally, these tuning-based methods introduce additional learning costs dependent on the pre-trained model and hinder the original distribution modeling ability of the base model.
Furthermore, in the TV2I task, mismatches between modalities can further hinder the model from accurately generating complex scenes due to the semantic gap between text and images and the complexity of multimodal data.
To this end, the ability to control the direction of information diffusion opens up the potential for a new branch of methodological paradigms to achieve versatile customization applications \emph{without the need to change the parameters of existing pre-trained diffusion models or learn any auxiliary neural networks}.

Following these insights, our preliminary work \emph{Cyclic One-Way Diffusion} (COW)~\cite{wang2023diffusion} achieves unidirectional diffusion for versatile customization scenarios, ranging from conventional visual-conditioned inpainting to visual-text-conditioned style transformation.
Methodologically, we re-inject the semantics (inverted latents) into the generation process and repeatedly “disturb” and “reconstruct” the image in a cyclic way. This training-free mechanism encourages information flow from the visual condition to the whole image while simultaneously minimizing reverse disruptive interference.
From the application point of view, the knowledge of the pre-trained diffusion model enables us to conduct meaningful editing or stylizing operations while maintaining the fidelity of the visual condition.

\begin{figure*}[t!]
    \centering
    \includegraphics[width=0.8\textwidth]{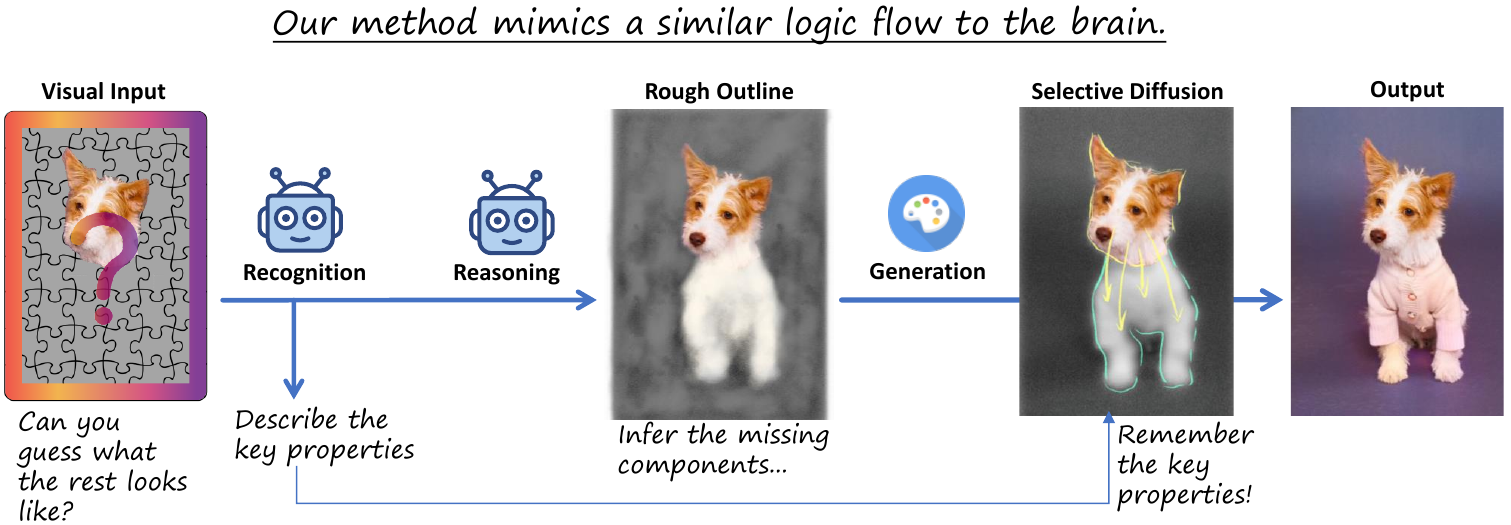}
    \caption{A cognitive-inspired approach for image generation. Starting with a partial visual input (left), we leverage a multimodal large language model to identify key properties (recognition) and infer the missing components (reasoning), guiding the generative model to complete the image (generation). Our selective diffusion mechanism further refines the process by directing information flow to the appropriate regions, ensuring the output (right) is contextually accurate and visually coherent.}
    \label{fig:brain}
\end{figure*}

Building on our previous work presented at ICLR 2024, COW~\cite{wang2023diffusion}, we present Selective One-Way Diffusion (SOW), a more comprehensive framework that refines the unidirectional information diffusion process to be more context-aware and precise in the following key areas. 
First, we leverage MLLMs to convert visual conditions into natural language descriptions, bridging the gap between textual and visual inputs and enhancing their consistency. MLLMs then analyze the relationships among the elements within the image to determine the optimal placement of these visual conditions and identify regions that closely interact with them. Next, drawing from insights provided by MLLMs, we implement dynamic attention modulation to control the direction and strength of information diffusion based on inter-regional correlations. This approach enables the model to intelligently allocate information, ensuring that diffusion is not only unidirectional but also contextually relevant, thereby maintaining spatial, semantic, and stylistic coherence throughout the generated image.
As shown in Fig.~\ref{fig:brain}, MLLMs conduct recognition and reasoning to identify key structures, while the generative model selectively fills in details based on this higher-level understanding. 
Extensive experiments and human studies, involving 1,200 responses for 600 groups of 512x512 images, demonstrate that SOW consistently outperforms its counterparts in terms of condition consistency and overall fidelity. Besides, SOW generates an image in just 5 seconds, far faster than other customization methods like DreamBooth~\cite{ruiz2022dreambooth}, which takes 732 seconds including image-specific training time.

\section{Related Work}\label{sec:related_work}
\label{sec:related0}

\textbf{Diffusion Models.}
The recent diffusion generative methods \cite{song2020ddim,2021ADM,nichol2021improved,song2020score,hoogeboom2021autoregressive,wu2022tune,zhu2023discrete} originate from the non-equilibrium statistical physics \cite{nonequilibrium}, which simulate the physical diffusion process by iteratively destroying the data structure through forward diffusion and restore it by a reverse annealed sampling process. DDPM~\cite{ho2020ddpm} shows the connection between stochastic gradient Langevin dynamics and denoising diffusion probabilistic models. DDIM~\cite{song2020ddim} generalizes the Markovian diffusion process to a non-Markovian diffusion process, and rewrites DDPM in an ODE form, introducing a method to inverse raw data into latent space with low information loss. In our work, we utilize DDIM inversion to access the latent space and find an information diffusion phenomenon during the sampling process towards the higher concentration data manifold. 

\textbf{Generative Dynamics and Regimes.}
Recent theoretical research provides insights into the dynamic behaviors of diffusion generative models, marking a promising direction in this field. Raya et al.~\cite{raya2024spontaneous} highlight how spontaneous symmetry breaking can enhance sample diversity. Additionally, Biroli et al.~\cite{biroli2024dynamical} provide a detailed theoretical framework that identifies three dynamical regimes within diffusion processes, demonstrating how speciation and collapse transitions are crucial in shaping the model's capabilities for memorization and generalization. 
In our approach, we explore the internal dynamics and information interactions within the denoising process of images, offering a new perspective for understanding and analyzing these phenomena. We hope this will encourage further research into the dynamics of diffusion models.

\textbf{Downstream Customization Generation.} 
Given a few images of a specific subject, the customization generation task aims to generate new images according to the text descriptions while keeping the subject's identity unchanged.
Early approaches mainly relied on GAN-based architectures~\cite{2016GAN-INT-CLS,2017StackGAN,2017SISGAN,2018StackGAN++,2018AttnGAN,2018ST,2019GP,2019StyleGANv1,2019Obj-GAN,2020StyleGANv2} for customization generation. 
In recent years, the diffusion methods under text condition image generation task (T2I) have made a great development \cite{nichol2021glide, ramesh2022hierarchical, midjourney, ramesh2021zero,ho2022classifier,saharia2022photorealistic}.
However, appointing a specific visual condition at a certain location on the results of a generation remains to be further explored. 
There are recent customized methods of learning-based visual-text conditioned generation like \cite{ruiz2022dreambooth, gal2022ti, controlnet, rombach2022ldm, dong2022dreamartist, brooks2023instructpix2pix, snoc, wei2023elite, guo2023animatediff, choi2021ilvr}. Methods like DreamBooth~\cite{ruiz2022dreambooth} and Textual Inversion~\cite{gal2022ti} learn the concept of the visual condition into a certain word embedding by additional training on the pre-trained T2I model \cite{kumari2022multi, dong2022dreamartist, ruiz2023hyperdreambooth, chen2023disenbooth}. It is worth noting that it is still hard for DreamBooth or Textual Inversion (TI) to keep the id-preservation even given 5 images as discussed in\cite{instantbooth}, and DreamBooth tends to overfit the limited fine-tuning data and incorrectly entangles object identity and spatial information, as discussed in \cite{PCAGen}. ControlNet~\cite{controlnet} trains an additional network for a specific kind of visual condition (e.g., canny edges, pose, and segmentation map).  DenseDiffusion\cite{kim2023dense} employs localized attention adjustments to achieve robust spatial control. More generally, an inpainting task is also a customization generation with a forced strong visual condition. Compared to these methods, our proposed method can explicitly preserve the pixel-level information of the visual conditions while achieving versatile application scenarios like style transfer and attribute editing.

\textbf{Enhancing Generative Performance with Large Models.}
    The recent advancements in Large Language Models (LLMs) have significantly revitalized the field of Natural Language Processing and the broader AI community, extending the functionality and reach of various applications. Leading models such as GPT-4~\cite{achiam2023gpt} and LLaMA~\cite{touvron2023llama,touvron2023llama2} are at the forefront of this progress, displaying exceptional skills in comprehension, reasoning, response generation, and maintaining expansive knowledge bases. 
    For the T2I task, approaches like~\cite{lian2023llm,wu2024self} utilize LLMs to provide a prior on the locations of complex textual information, thereby controlling the spatial generation of diffusion models.
    Building on these advancements, the research community has redirected its efforts toward the development of Multimodal Large Language Models (MLLMs)~\cite{team2023gemini,driess2023palm,zhu2023minigpt}. These models are specifically designed to equip LLMs with the capability to process and interpret both images and text, thereby broadening their applicability across various modalities. Notably, certain MLLMs have distinguished themselves through exceptional visual reasoning capabilities, with Gemini~\cite{team2023gemini} standing out as a particularly competitive model in this domain. In our work on the TV2I  task, we harness the capabilities of the MLLM to markedly improve image generation performance.

\begin{figure*}[!t]
    \centering
    \includegraphics[width = 0.6\textwidth]{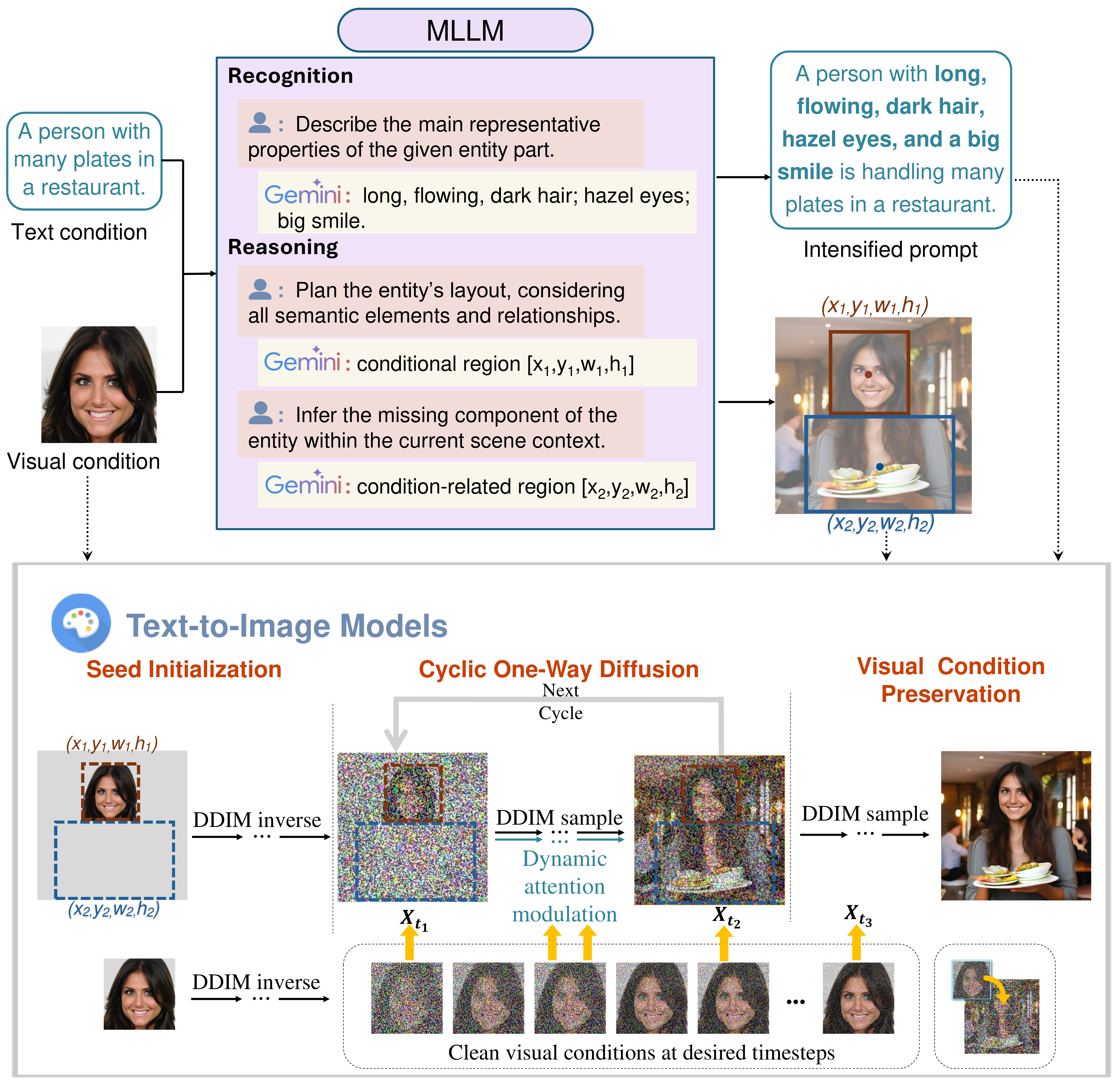}
    \caption{The pipeline of our proposed SOW method. Initially, given the visual condition and text condition, we employ a MLLM  Gemini~\cite{team2023gemini} to infer the textual description, the adaptive location box of the visual conditional region, and the box of the condition-related region through a three-stage reasoning process. The input visual condition is then affixed to a predefined background, serving as the seed initialization for the cycle. During the Cyclic One-Way Diffusion process, we ``disturb'' and ``reconstruct'' the image in a cyclic way and ensure a continuous one-way diffusion by consistently replacing the image with corresponding $\mathbf{x_t}$. Also, by integrating these prior pieces of information, we execute cyclic diffusion with dynamic attention modulation, enhancing the coherence and accuracy of the generated outputs.} 
    \label{fig:Method}
\end{figure*}

\begin{figure*}[t]
    \centering
    \includegraphics[width=0.9\textwidth]{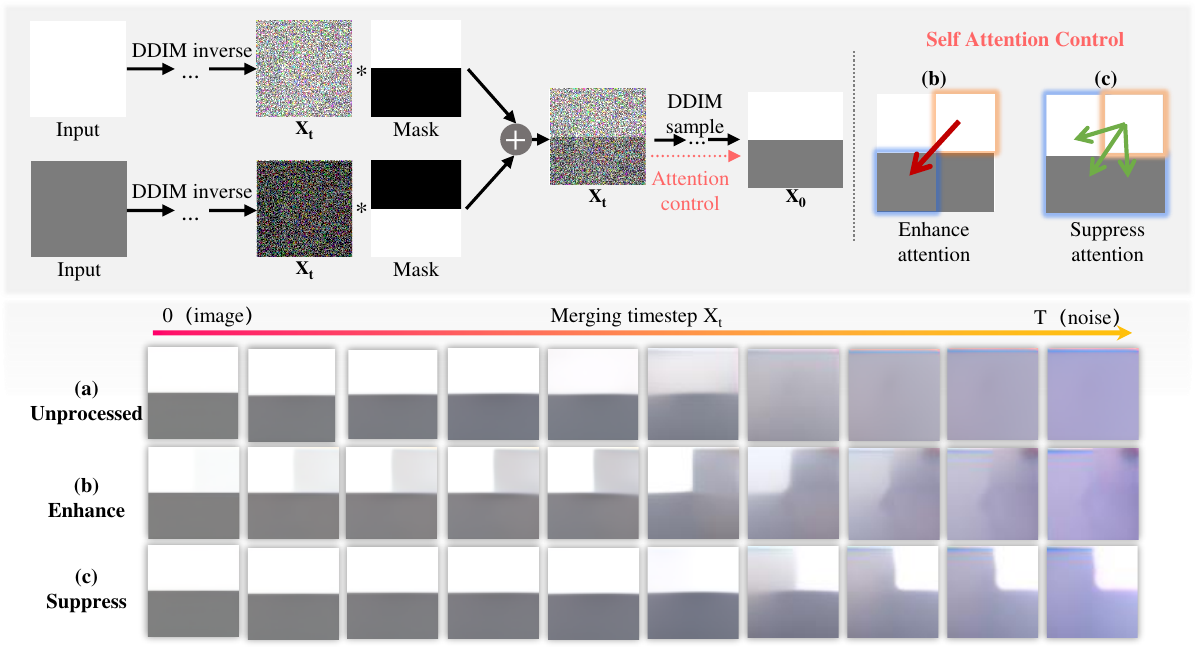}
    \caption{
    Illustration of \emph{``diffusion in diffusion''}. In experiment (a), We invert the pictures of pure gray and white to $\mathbf{x_t}$, merge them together, and then regenerate them to $\mathbf{x_0}$ via deterministic denoising. In experiment (b), we enhance the attention scores of the upper right quartile to the lower left quartile, while in experiment (c) we suppress attention scores from the upper right quartile towards other areas. The resulting images show how regions within an image diffuse and interfere with each other during denoising, and reveal the direct effect of attention on diffusion.}
    \label{fig:diffuse_in_diffusion}
\end{figure*}

\section{Generative Dynamics in Diffusion Models}
In this section, we present the preliminary knowledge of diffusion models, followed by an in-depth investigation into their generative dynamics and regimes from a fresh perspective.
\subsection{Preliminaries}\label{subsec:Preliminaries}
Denoising Diffusion Probabilistic Models (DDPMs)~\cite{ho2020ddpm} define a Markov chain to model the stochastic random walk between the noisy Gaussian space and the data space, with the diffusion direction written as,
\begin{equation}\label{eq:diffusion_iter}
    q(\mathbf{x}_t|\mathbf{x}_{t-1})=\mathcal{N}(\sqrt{1-\beta_t}\mathbf{x}_{t-1}, \beta_t I),
\end{equation}
where $t$ represents diffusion step, $\{\beta_t\}^T_t$ are usually scheduled variance, and $\mathcal{N}$ represents Gaussian distribution. 
Then, a special property brought by Eq.~\ref{eq:diffusion_iter} is that:
\begin{equation}\label{eq:add_noise}
    q(\mathbf{x}_t|\mathbf{x}_{t-k})=\mathcal{N}(\sqrt{\alpha_t/\alpha_{t-k}}\mathbf{x}_{t-k}, \sqrt{1-\alpha_t/\alpha_{t-k}} I),
\end{equation}
where $\alpha_t=\prod \limits_{i=0}^t ( 1-\beta_i)$. So we can bring $\mathbf{x}_{t}$ to any $\mathbf{x}_{t+k}$ in a one-step non-Markov way in our proposed cyclic one-way diffusion (Sec.~\ref{sec:Cyclic}) by adding certain random Gaussian noise.

DDIM~\cite{song2020ddim} generalizes DDPM to a non-Markovian diffusion process and connects the sampling process to the neural ODE: 
\begin{equation}\label{eq:ode}
    d\bar{\mathbf{x}}(t) = \epsilon_\theta^{(t)}(\frac{\bar{\mathbf{x}}(t)}{\sqrt{\sigma^2+1}})d\sigma(t).
\end{equation}
By solving this ODE using Euler integration, we can inverse the real image $\mathbf{x}_0$ (the visual condition) to $\mathbf{x}_t$ in any corresponding latent space $\epsilon_\theta^{(t)}$~\cite{zhu2023boundary, mokady2023null, asperti2023head} while preserving its information. The symbols $\sigma$ and $\bar{\mathbf{x}}$ are the reparameterizations of $(\sqrt{1- \alpha}/\sqrt{\alpha})$ and $(\mathbf{x}/\sqrt{\alpha})$ respectively.

\subsection{Diffusion in Diffusion}
\label{subsec:diffusion_in_diffusion}

\textbf{Internal Interference in Diffusion Generation.} 
Diffusion in physics is a phenomenon caused by random movements and collisions between particles.
The diffusion model, drawing inspiration from non-equilibrium thermodynamics, establishes a Markov chain between a target data distribution and the Gaussian distribution.
Subsequently, it learns to reverse this diffusion process, thereby constructing the desired data samples from the Gaussian noise. 
This inherently simulates a gradual, evolving process that can be viewed as a random walk through a large number of possible data distributions, which eventually gradually approaches the real data distribution. Therefore, diffusion models share an interference phenomenon similar to that of physical diffusion, characterized by continuous information exchange within the data, ultimately achieving harmonious generation results.

As illustrated in Fig.~\ref{fig:diffuse_in_diffusion}, we design a toy experiment to reveal this phenomenon more intuitively. To start with, we apply DDIM Inversion to convert gray and white images into various latent codes along the diffusion timeline, spanning from the start ($t=T$) to the final state ($t=0$). Existing literature \cite{song2020ddim, zhu2023boundary}  demonstrates that those intermediate latent codes can well reconstruct the raw image via deterministic denoising. In other words, both latent codes contain information inherited from their respective raw images, i.e., pure gray and white colors.
Consequently, at each selected time step $t$, we merge half of the latent codes from the two images into one and denoise the resulting combination.
This allows us to observe how different pieces of information interact throughout the generation process and thus influence the final image.
The result (a) in Fig.~\ref{fig:diffuse_in_diffusion} shows that as the merging time step $t$ (in which step we put two latent codes together) approaches $T$ (the Gaussian noise end), the corresponding denoised image $x_0$ exhibits the spatial diffusion phenomenon, resulting in stronger color blending. Conversely, as $t$ approaches $0$ (the raw image end), the image showcases robust reconstruction ability, with minimal interference between the two colors.
This experiment demonstrates the existence of mutual interference throughout the generation process and highlights its varying intensity across different stages of denoising.

\textbf{Varying Interference Intensity in Diffusion Generation.}
Based on previous research on diffusion phases~\cite{biroli2024dynamical,raya2024spontaneous} and our supplementary experiments in Appendix~\ref{sec:properties}, we can roughly divide the denoising process into three distinct stages. Throughout the reverse diffusion process, the model attends to different levels of information at each stage, essentially embodying a progression from extreme noise to semantic formation to refinement.
Introducing guidance too early hinders its reflection in the final image due to the uncontrollable interference caused by excessive noise while introducing it too late does not allow for the desired high-level semantic modifications.
The best way to inject visual condition information is in the middle stage, where the model gains the capacity to comprehend and generate basic semantic content, striking a balance between controllable inner mutual influence and responsiveness to text conditions.
Ultimately, proper refinement at the last stage ensures that the generated images exhibit intricate details of visual condition.
These insights and observations pave the way for integrating new visual condition control paradigms into pre-trained diffusion models.

\textbf{Attention-guided Information Diffusion.}
The interference observed during diffusion generation stems not only from the unique iterative denoising theory but also from the model's architecture (e.g., the potential inductive bias of convolution and self-attention mechanism). 
Specifically, convolution layers, with their local receptive fields, capture fine-grained dependencies between neighboring pixels, while the self-attention mechanism enables the model to establish global interactions. 
This combination facilitates information diffusion across the entire image, which, though necessary for coherence, can lead to unwanted blending in certain tasks.
However, by adjusting the self-attention scores, we can dynamically control how much information is exchanged between different regions of the image.
In our toy experiment, as shown in Fig.~\ref{fig:diffuse_in_diffusion}, increasing attention between regions intensifies diffusion (experiment (b)), while reducing attention preserves local characteristics, minimizing blending (experiment (c)).
These findings demonstrate that attention modulation acts as a powerful steering mechanism for the diffusion of information. Through fine-tuning attention scores, we can gain precise control over the speed, direction, and intensity of information diffusion within the image.

\section{Selective One-Way Diffusion}\label{method}
\label{sec:method}
In this section, we introduce our proposed method \emph{Selective One-Way Diffusion} for the TV2I task. As illustrated in Fig.~\ref{fig:Method}, We first introduce MLLM-driven stages that intensify prompts and provide conditions for specific regions (Sec.~\ref{sec:mllm}) to address mismatches between the two modalities and enhance control over the generative model. Subsequently, the outputs—namely the conditional region, condition-related region, and intensified prompt—are fed into the foundational \emph{Cyclic One-Way Diffusion (COW)} framework (Sec.~\ref{sec:Cyclic}). During the COW process, dynamic attention modulation is applied (Sec.~\ref{sec:AM}), where we utilize region information provided by the MLLM to control the internal diffusion direction, thus enhancing the coherence of image generation. For clarity, a pseudo-code implementation is provided in Algo.~\ref{algorithm}.

\begin{algorithm}[t]
\caption{Selective One-Way Diffusion}
\label{algorithm}
\textbf{Input:} Visual condition $V_0$, text prompt $c$, target cycle number $N$, ends of the cycle $t_1, t_2$, ID preserve step $t_3$ \\
\textbf{Output:} Generated Image $x$ \\
\textbf{Definition:} Deterministic sampling $p_{d}$, Deterministic sampling with guided attention $p_{d,a}$, Stochastic sampling $p_s$, Indicator $R_{c}$, Threshold $\tau$, Visual condition region $\text{box}_{v}$, Condition-related region $\text{box}_{r}$, Intensified prompt $c'$
\SetKwFunction{Encoder}{Encoder}
\SetKwFunction{Gemini}{Gemini}
\SetKwFunction{ODEInverse}{ODEInverse}
\SetKwFunction{InjectNoise}{InjectNoise}
\SetKwFunction{Paste}{Paste}
\SetKwFunction{Replace}{Replace}
\begin{algorithmic}[1] %[1] enables line numbers
\STATE \textbf{//Step 1: Model Loading and Initialization} % Display Stage 1
\STATE $\text{box}_{v},\text{box}_{r}, c^\prime \gets \Gemini(V_0, c)$
\STATE $X_0 \gets$ Paste $V_0$ into $\text{box}_{v}$
\STATE $x_0, v_0 \gets \Encoder(X_0), \Encoder(V_0)$
\STATE $x_{t_2}, v_t \gets \ODEInverse(x_0, t_2), \ODEInverse(v_0, t_2)$

\STATE \textbf{//Step 2: Cyclic One-Way Diffusion with Dynamic Attention Modulation} % Display Step 2
\WHILE{$\text{cyc}_{num} < N$}
    % \FOR{$t = t_1, t_1-1,\ldots,t_2$ do}
    \FOR{$t = t_2, \ldots,t_1+1,t_1$ do}
        \STATE $x_t' \gets \Replace(x_t, v_t)$
        \STATE Calculate $R_{\text{c}}$ (Eq.~\ref{eq:r_c})
        \IF{$R_{\text{c}} > \tau$}
            \STATE $x_{t-1} \gets p_{d,a}(x_t', t, c',\text{box}_{r})$ (Eq.~\ref{eq:attn_mod})
        \ELSE 
            \STATE  $x_{t-1} \gets p_{d}(x_t', t, c')$
        \ENDIF
    \ENDFOR
    \STATE $x_{t_2} \gets \InjectNoise(x_{t_1})$
\ENDWHILE
\STATE \textbf{//Step 3: Visual Condition Preservation} % Display Step 3
\FOR{$t = t_1,t_1-1,\ldots,0$ do}
    \IF{$t == t_3$}
        \STATE $x_t' \gets \text{Replace}(x_t, v_t)$
        \STATE $x_{t-1} \gets p_s(x_t', t, c)$
    \ENDIF
    \STATE $x_{t-1} \gets p_s(x_t, t, c)$
\ENDFOR
\end{algorithmic}
\end{algorithm}

\subsection{Enhancing TV2I Generation with MLLM-driven Stages}\label{sec:mllm}

In the TV2I task, the model faces the significant challenge of understanding the semantic information in the textual description and translating this information into visually specific details.
To refine the subsequent generative process and address the complex context-driven challenges, we introduce a series of stages leveraging Multimodal Large Language Models (MLLMs).
In this work, we exploit Gemini~\cite{team2023gemini}, a multi-modal language model that accommodates various data formats like images, text, and audio.
By harnessing Gemini's advanced contextual understanding and reasoning, we aim to enhance the alignment between text and visual modalities, optimize spatial layouts, and ensure coherent information flow.

\textbf{Stage 1: Prompt Intensification.}  
In this stage, we focus on \emph{Recognition}, where Gemini is tasked with generating descriptions capturing salient visual features from the given visual condition. Subsequently, we fuse the generated textual description with the user’s textual input, creating an intensified prompt \(c^\prime\). 
This refined prompt serves as input for the T2I model, providing a more comprehensive representation of both visual and textual information.
As a result, it addresses the inherent global-local misalignment and abstraction differences between text and visual modalities in the TV2I task, improving the model’s ability to generate images that accurately reflect the intended context.

\textbf{Stage 2: Adaptive Position.}\label{para:stage2}
In parallel with Stage 1, we engage in \emph{Reasoning} to analyze spatial relationships and element placements. Gemini assesses both the visual and textual conditions, considering all relevant semantic elements. It generates bounding boxes for each visual condition, specified in [x, y, width, height] format where x and y represent the coordinates of the top-left corner of the region. These bounding boxes serve as precise spatial guidelines for seed initialization (Sec.~\ref{sec:Cyclic}), ensuring that visual conditions are placed in a sensible and aesthetically pleasing manner. 
This step facilitates a harmonious arrangement of visual elements, contributing to an overall improved composition in the generated images.
    
\textbf{Stage 3: Contextual Refinement.}
Building on the spatial arrangement from Stage 2, we further apply \emph{Reasoning} to refine the contextual information.
Gemini identifies condition-related regions that directly interact with the visual condition, mitigating information leakage and preventing truncated object outlines. For instance, when considering a face as the visual condition, Gemini recognizes that the body beneath is directly relevant, while the surrounding background is deemed irrelevant.
This step is crucial for ensuring that the generated images maintain semantic coherence, such as preventing disconnection between a person's head and body. 
To achieve this, the attention modulation mechanism (Sec.~\ref{sec:AM}) integrates this understanding by selectively directing key information flows, ensuring that relevant details (e.g., facial features) reach the appropriate regions (e.g., the body).
This targeted approach helps maintain a coherent and contextually relevant output, avoiding issues where elements appear fragmented or misaligned.
In summary, these MLLM-driven stages collectively enhance the generative model’s performance, ensuring that the output achieves high semantic coherence, accurate spatial relationships, and a refined integration of visual and textual conditions.

\subsection{Training-free Cyclic One-Way Diffusion Framework}\label{sec:Cyclic}
To address the chaotic interference inherent in disordered diffusion, we introduce Cyclic One-Way Diffusion (COW). COW restructures the diffusion process into a unidirectional flow, effectively minimizing disruptive interference and ensuring efficient information transfer. This framework forms the core of our method, enabling versatile and efficient pixel-level and semantic-level visual conditioning without training.
COW comprises three key components: Seed Initialization, Cyclic One-Way Diffusion Process, and Visual Condition Preservation, as shown in Fig.~\ref{fig:Method}.

\textbf{Seed Initialization.} 
The objective of this mechanism is to inject stable high-level semantic information early in the denoising process, effectively reducing the layout conflicts with the visual condition. 
Classical random initialization sampled from the Gaussian distribution of diffusion model may introduce features or structures that conflict with the given visual condition.  
For instance, if an object is intended to appear on the left but the initial noise favors the right, there will be a conflict. 
The model must invest considerable effort during generation to correct this inconsistency, which may still negatively impact the quality of the generated images.
To avoid such conflicts, we introduce a novel initialization: embedding visual conditions directly onto a predefined background, usually a semantically neutral pure gray.
This process uses adaptive positioning, which leverages semantic cues from multimodal large language models (MLLMs) to determine optimal object placement (discussed further in Sec.~\ref{sec:mllm}).

\textbf{Cyclic One-Way Diffusion Process.}
Practically, we invert the visual condition into its latent representation by solving the probabilistic flow ODE (Eq.~\ref{eq:ode}) and embedding it in the initial random Gaussian noise that serves as the starting point, which can provide a good generative prior to maintain consistency with the visual condition. 
However, the implanted information will be continuously disrupted by inner diffusion from the surrounding Gaussian region at every denoising step according to the analysis of Sec.~\ref{subsec:diffusion_in_diffusion}.
Therefore, we introduced ``one-way" and ``cyclic" strategies to maximize the flow of information from the visual condition to the whole image and minimize undesired interference from other image regions.
%one-way
To be specific, we store inverted latents of the visual condition at each inversion step in the middle stage (the semantic formation stage), denoted as ${\mathbf{x}_{t_{1}}, \mathbf{x}_{t_{1}+1}, \ldots, \mathbf{x}_{t_{2}}}$ and gradually embed them in the corresponding timesteps during the generation process. Through this step-wise information injection, we can ensure the unidirectional propagation of information, i.e., it only propagates from the visual condition to the other regions without interference from information in the background or other parts of the image. 
%cycle
Given the limited generative capacity of the model at each step, noise is injected to regress the generative process to earlier stages, as illustrated in Eq.~\ref{eq:diffusion_iter}. This cyclic utilization of the model's generative capacity enables the continuous perturbation of inconsistent semantic information, facilitating the re-diffusion of conditional guidance in subsequent rounds. To provide the model with greater exploratory space, we have increased the degree of perturbation by injecting noise at a larger scale in a backward step.
The cyclic one-way process benefits the model from one-way guidance from the visual condition, creates additional space by cycles for semantic “disturb” and “reconstruct”, and ultimately achieves harmony among the background, visual condition, and text condition.

\textbf{Visual Condition Preservation.}
Conflicts between the visual and the text conditions often exist (such as a smiling face condition and a ``sad'' text prompt), necessitating a method that can effectively balance these conditions. We observe that the middle stage is still subject to some extent of uncertainty, which in turn leaves enough space for controlling the text condition guidance on the generation of the visual condition region. Meanwhile, in the later stage, the model focuses on refining high-frequency details and textures to enhance image quality while maintaining global structure integrity. Thus we explicitly control the degree of visual condition preservation by replacing the corresponding region at an adjustable step $\mathbf{x_{t_3}}$ in the early phase of this later stage. This approach can effectively preserve fidelity to both the visual and the text conditions, achieving harmonious results of style transfer and attribute editing without additional training.

\subsection{Dynamic Attention Modulation}\label{sec:AM}

While COW effectively manages the unidirectional information flow from visual conditions, it may still face challenges such as over-penetration of target information into irrelevant regions or insufficient guidance to relevant regions, potentially resulting in artifacts like truncated outputs. 
For example, in the attribute editing example in the first column of the second row of Fig.~\ref{fig:Ablation_comparison}, when the face serves as the visual condition, there is no cohesive integration around it. This results in a truncated output that disconnects from the background.
% informed by contextual insights from MLLMs, out approach enhances both the accuracy and coherence of generated images.
These issues stem from the intricate, context-driven relationships that are not always fully captured by current generative models. To address the aforementioned issues, SOW employs dynamic attention modulation inspired by DenseDiffusion~\cite{kim2023dense} to refine and precisely control the flow of information. Drawing on contextual insights from Multi-Level Language Models (MLLMs), our approach enhances both the accuracy and coherence of the generated images.

\begin{figure}[t]
\centering
\includegraphics[width = 0.45\textwidth]{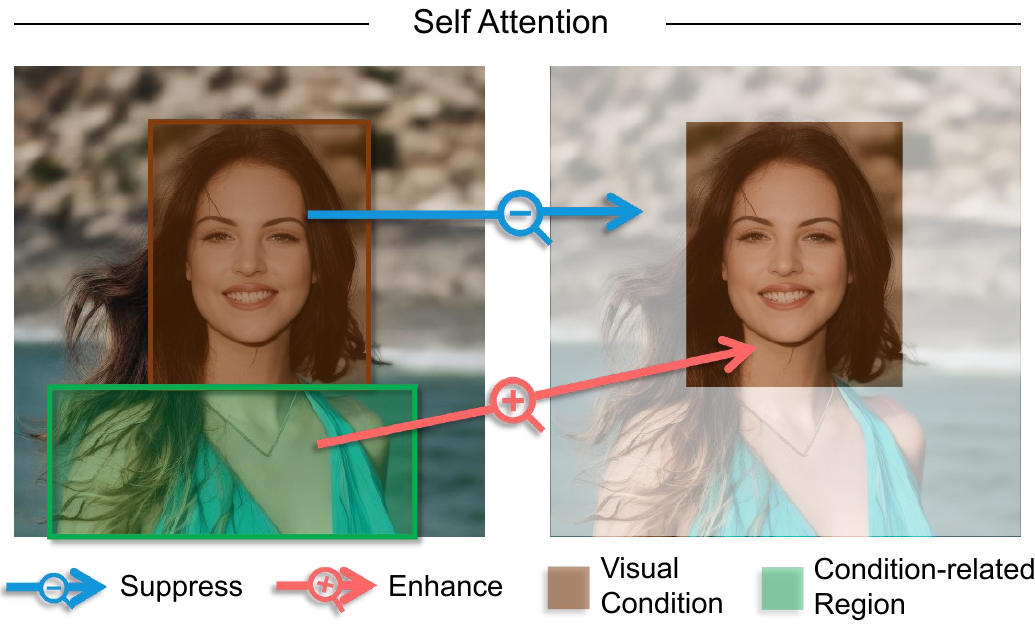}
\caption{Schematic of attentional modulation. We modulate the attention dynamically by suppressing the score originating from the face region (highlighted by a brown frame, termed the conditional region)  to areas \emph{outside} the face region (termed non-conditional regions). Simultaneously, we enhance the attention score from the body region (marked with a green frame and termed condition-related region) towards the face region, to better facilitate consistency.}
\label{fig: attention_control}
\end{figure}

As illustrated in Fig.~\ref{fig: attention_control}, the general idea is to:
\begin{itemize}
\item Suppress the attention from the conditional region to non-conditional regions (denoted as $P^{-}$) to preserve the integrity of the conditional information.
\item Enhance the attention from condition-related regions to the conditional region (denoted as $P^{+}$) to direct a more effective and targeted information flow.
\end{itemize}

The attention control function is defined as:
\begin{equation}  
P^{-} =  - \mathcal{M}_{c \to nc} \cdot \gamma^{\text{time}}\left(t\right) \cdot \omega^{-}, 
\end{equation}
\begin{equation}  
P^{+} =  \mathcal{M}_{cr \to c} \cdot \gamma^{\text{dis}}\left(D_{cr \to o^*}\right) \cdot \gamma^{\text{time}}\left(t\right) \cdot \omega^{+},   
\end{equation}
where $\mathcal{M}_{c \to nc}$ denotes the attention mask from the conditional region to the non-conditional region, and $\mathcal{M}_{cr \to c}$ represents the attention mask from the condition-related region to the conditional region. By leveraging the robust reasoning capability of MLLMs, we can employ MLLMs for spatial planning, thereby obtaining the expected condition-relevant regions simultaneously (See Sec.~\ref{sec:mllm}). Moreover, where $D_{cr \to o*}$ denotes the distance from condition-related regions to the conditional center $o^{*}$.
$\omega^{-}$ and $\omega^{+}$ are adjustment factors, $t$ is the current cycle count, and $\gamma(x)$ is a decay function that adjusts attention over time and distance:  
\begin{equation}  
\gamma(x) = \left(\frac{1}{2} \cdot \left(1 + \cos\left(\pi \cdot \frac{2T - x}{T}\right)\right)\right)^{a},  
\end{equation}  
where $T$ and $a$ are predefined parameters that control the rate and extent of decay.
This ensures that as the image generation progresses, additional guidance diminishes, focusing more on regions close to the designated region. This smooth transition maintains the stability and coherence of information flow, preventing fragmented outputs.

However, it is unreasonable to uniformly increase the attention scores for all images, as those that have already achieved good generation quality through the original COW process might experience a degradation in generation performance if their attention scores are overly augmented or diminished. 
An indicator \(R_{\text{c}}\) is computed to evaluate the distribution and extent of information diffusion from the conditional region to the condition-related region. Specifically, \(R_{\text{c}}\) measures whether the sum of attention scores for the closest one-fifth of points within the guiding box to the center of the visual condition occupies a significant portion of the guiding box as follows:
\begin{equation}
R_{\text{c}} = \frac{1}{H} \sum_{h=1}^{H} \left( \frac{\sum_{j=1}^{k} a_{vg, j}^{(h)}}{\sum_{i=1}^{m} a_{vg,i}^{(h)}} \right),
\label{eq:r_c}
\end{equation}
where $k = \lceil \frac{m}{5} \rceil$ and $m$ is the total number of points in the guiding box, and $H$ represents the number of attention heads. Let $A = \frac{Q \cdot K^T}{\sqrt{d_k}} $ denote the logits before applying the softmax function, $a$ represents attention logits for each point. The degree of attention modulation towards the visual condition is then adjusted based on the difference between this sum and a predetermined threshold denoted.
Finally, the self-attention score is updated by:
\begin{equation}
A^{\prime} = A +  \sqrt{\max(0, R_{c} - \tau)} \cdot (P^{+} + P^{-}).
\label{eq:attn_mod}
\end{equation}

Before each operation, we perform min-max normalization on the scores. By dynamically adjusting the attention allocation based on the evolving generation process, the model can adapt in real-time to optimize information dissemination. This approach not only improves the accuracy of COW but also enhances its robustness and adaptability in handling a wide range of visual scenarios.

\section{Experiments}\label{sec:experiments}

\begin{table*}[!t]
  \caption{Quantitative results of objective metrics on the CelebA-TV2I test set and human evaluations.
  }
  \vspace{1mm}
  \centering
  \setlength{\aboverulesep}{0pt}
  \setlength{\belowrulesep}{0pt}
  \scalebox{1}{
  \begin{tabular}{l|ccc|cc}
    \bottomrule
    % \bottomrule
    & \multicolumn{3}{c|}{Objective Metrics}& \multicolumn{2}{c}{Human Evaluations}   \\ \hline
    Methodology  &ID-Distance $\downarrow$ & Face Detection Rate $\uparrow$ & Time Cost $\downarrow$ & Condition Consistency $\uparrow$ & General Fidelity $\uparrow$ \\ \hline
    TI'22~\cite{gal2022ti}        & 1.201 & 44.50\% & 3025s &  3.43\%     & 10.41\%  \\
    DreamBooth'22~\cite{ruiz2022dreambooth}    & 1.326 & 38.67\% & 732s  & \underline{25.04\%}    & \underline{32.26\%}      \\
    ControlNet'23~\cite{controlnet}     & 1.092 & \underline{96.16\%} & \textbf{4s}  & 9.31\%    & 3.26\% \\
    SD inpainting'22~\cite{rombach2022ldm}  & \textbf{0.408} & \textbf{100.00\%} & \underline{5s} & 7.21\%     & 1.25\%  \\
    SOW (ours)      & \underline{0.771} & \textbf{100.00\%} & \underline{5s} &\textbf{55.00\%}   & \textbf{52.83\%} \\
    \bottomrule
  \end{tabular}}
  \vspace{-1mm}
  \label{Tab:metric-table}
\end{table*}

\textbf{Benchmark.}
To simulate the visual condition processing in real scenarios, we adopt face images from CelebAMask-HQ ~\cite{lee2020maskgan} as our visual condition. We design three kinds of settings for the text conditions\textemdash normal prompt, style transfer, and attribute editing. Subsequently, we pair each prompt with two images, compiling these image-text pairs into the CelebA-TV2I dataset, which serves as the conditions for the TV2I task.

\textbf{Baselines.} We perform a comparison with four existing works incorporating different levels of the visual condition into pre-trained T2I models: DreamBooth\cite{ruiz2022dreambooth} based on the code\footnote{https://github.com/ShivamShrirao/diffusers/tree/main/examples/dreambooth}, TI~\cite{gal2022ti}, SD inpainting~\cite{rombach2022ldm}, and ControlNet on the canny edge condition~\cite{controlnet}.
DreamBooth~\cite{ruiz2022dreambooth} introduces a rare token identifier along with a class prior for a more specific few-shot visual concept by fine-tuning pre-trained T2I model to obtain the ability to generate specific objects in results.
TI~\cite{gal2022ti} proposes to convert visual concept to word embedding space by training a new word embedding using a given visual condition, and uses it directly in the generation of results of specific objects.
ControlNet~\cite{controlnet} incorporate additional visual conditions (e.g., canny edges) by finetuning a pre-trained Stable Diffusion model in a relatively small dataset (less than 50k) in an end-to-end way.
SD inpainting~\cite{rombach2022ldm} preserves the exact pixel of the visual condition when generating an image, and it is the inpainting mode of the pre-trained Stable Diffusion.

\textbf{Implementation Details.} We use a single NVIDIA 4090 GPU to run experiments since our proposed SOW method is training-free. We implement SOW, SD inpainting, and ControlNet on pre-trained T2I Stable Diffusion model~\cite{rombach2022ldm} sd-v2-1-base, with default configuration condition scale set 7.5, noise level $\eta$ set 1, image size set 512, 50 steps generation (10 steps for SOW), and negative prompt set to ``a bad quality and low-resolution image, extra fingers, deformed hands". Note that we implement DB and TI following their official code and use the highest supported Stable Diffusion version sd-v1-5. 
 
We choose $\mathbf{x_{t_1}}$ to step 5, $\mathbf{x_{t_2}}$ to 7, cycle number to 10. We use slightly different settings for the three different tasks. We set $\mathbf{x_{t_3}}$ to be [4, 3], eta to be 0 in the normal prompts, $\mathbf{x_{t_3}}$ to be 4, eta to be 0.1 in the attribute editing prompts, and $\mathbf{x_{t_3}}$ to be 4, eta to be 1 in the style transfer prompts. 
Additionally, we employ the Gemini-1.5-pro version of the Gemini\cite{team2023gemini} when performing MLLM-driven stages and restrict the MLLM's predictions for the size of the conditional region to between 180 and 256. For \(P^-\), we set \(\omega^- = 0.04\), $T = 10$ and \(a = 0.5\) in \(\gamma^\text{time}\); For \(P^+\),we set \(\omega^+ = 0.145\) and \(a = 0.8\) in \(\gamma^\text{time}\) , with $T = 10$ and \(a = 0.3\) in $\gamma^\text{dis}$. Throughout the entire process, we choose \(\tau = 0.273\). For analysis, we partitioned the dataset, allocating 10\% as a validation set for tuning hyperparameters and the ablation studies. The remainder was used as a test set for comparative analysis against other baselines.
\begin{table}[!t]
\setlength{\tabcolsep}{4pt}
\caption{Ablation study of the main improvements in Selective One-Way Diffusion (SOW) on the CelebA-TV2I validation set: adaptive position (AP), dynamic attention modulation (DAM), and prompt intensification (PI). The face detection rate remains 100\% in all four experiments. }

    \vspace{1mm} e
    \centering
    \setlength{\aboverulesep}{0pt}
    \setlength{\belowrulesep}{0pt} 

    \scalebox{1}{
    \begin{tabular}{ccc|cc}
            \toprule

         AP    & DAM   & PI   & ID-Distance $\downarrow$   & Failure Rate $\downarrow$  \\  \hline \hline
   
        \xmark & \xmark & \xmark & 0.8515 & 16.67\% \\
        \cmark & \xmark & \xmark   & 0.8239    & 13.33\%\\

        \cmark & \cmark & \xmark & 0.8089    & 8.33\% \\
        \cmark & \cmark & \cmark  & \textbf{0.8068}  & \textbf{5.00\%}  \\
        \bottomrule
    \end{tabular}}
    \vspace{-1mm} % 
    \label{tab:ablation_o}
\end{table}

\begin{table}[!t]
\setlength{\tabcolsep}{6pt}
    \caption{
    Ablation study of Dynamic Attention Modulation on the CelebA-TV2I validation set. The face detection rate remains 100\% in all six experiments.
    }
    \vspace{1mm} 
    \centering
    \setlength{\aboverulesep}{0pt} 
    \setlength{\belowrulesep}{0pt} 
    \centering
    \scalebox{1}{
    \begin{tabular}{cc:ccc|ccccc}
            \toprule
        % & & & & a & b\\
         $P^-$ & \multicolumn{1}{c}{$P^+$} & $\gamma^{time}$ & $\gamma^{dis}$ & $R_{\text{c}}$   & ID-Distance $\downarrow$ & Failure Rate $\downarrow$ \\ \hline \hline
        % \midrule
        \xmark & \cmark & \cmark        & \cmark           & \cmark       &    0.8162          & 11.67\%       \\	

        \cmark & \xmark & \cmark        & \cmark           & \cmark    &   0.8984      & 16.67\%              \\	

        \cmark & \cmark & \xmark        & \cmark           & \cmark    &    0.8594      & 6.67\%            \\

        \cmark & \cmark & \cmark        & \xmark           & \cmark      & 0.8214          & 13.33\%           \\

        \cmark & \cmark & \cmark        & \cmark           & \xmark         &  0.8103          &8.33\%             \\
        \cmark & \cmark & \cmark        &\cmark             &\cmark     & \textbf{0.8068}   &\textbf{5.00\%}   \\
    
        \bottomrule
    \end{tabular}}
    \vspace{-1mm} 
    \label{tab:ablation_r}
\end{table}

\subsection{Comparisons to the Existing Methods}
\textbf{Evaluation Metrics.}
To evaluate the quality of the generated results for this new task, we first consider the assessment of visual fidelity. We adopt a face detection model (MTCNN~\cite{zhang2016joint}) to detect the presence of the face, and a face recognition model (FaceNet~\cite{schroff2015facenet}) to obtain the face feature and thus calculate the face feature distance between the generated face region and the given visual condition as the Face Detection Rate and ID-Distance metric.
However, relying solely on model predictions can not fully capture the subtle differences between images and can not reflect the overall quality of the images (e.g., realism, richness), which are critically crucial for human perception. 
Therefore, we further evaluate our model via \textbf{human evaluation}. We design two base criteria and invited 50 participants to be involved in this human evaluation. The two criteria are: 1. Condition Consistency: whether the generated image well matches the visual and textual conditions; 2. General fidelity: whether the chosen image looks more like a real image in terms of image richness, face naturalness, and overall image fidelity. It's important to note that when assessing the latter criterion, participants are not provided with textual and visual conditions to prevent additional information from potentially interfering with the assessment process.

\textbf{Qualitative Results.} 
As shown in Fig.~\ref{fig:teaser}, our method consistently demonstrates superior fidelity to both textual and visual conditions, as well as overall image naturalness across all three TV2I task settings. In contrast, other learning-based methods struggle to maintain performance across various scenarios. Specifically, TI~\cite{gal2022ti} and DreamBooth~\cite{ruiz2022dreambooth} encounter difficulties in preserving face identity features and generating valid face-containing images. ControlNet~\cite{controlnet} produces cut-off outputs due to the mismatch between the provided local face conditions and the model's learned global control patterns. While Stable Diffusion inpainting mode~\cite{rombach2022ldm} enforces the preservation of visual conditions, it struggles with large (75\%) completion areas. Furthermore, this rigid preservation limits its flexibility in tasks like attribute editing and style transfer.
More comparisons of generated images are included in Fig.~\ref{compare}.
\begin{figure}[t]
  \centering
 
  \includegraphics[width=0.5\textwidth]{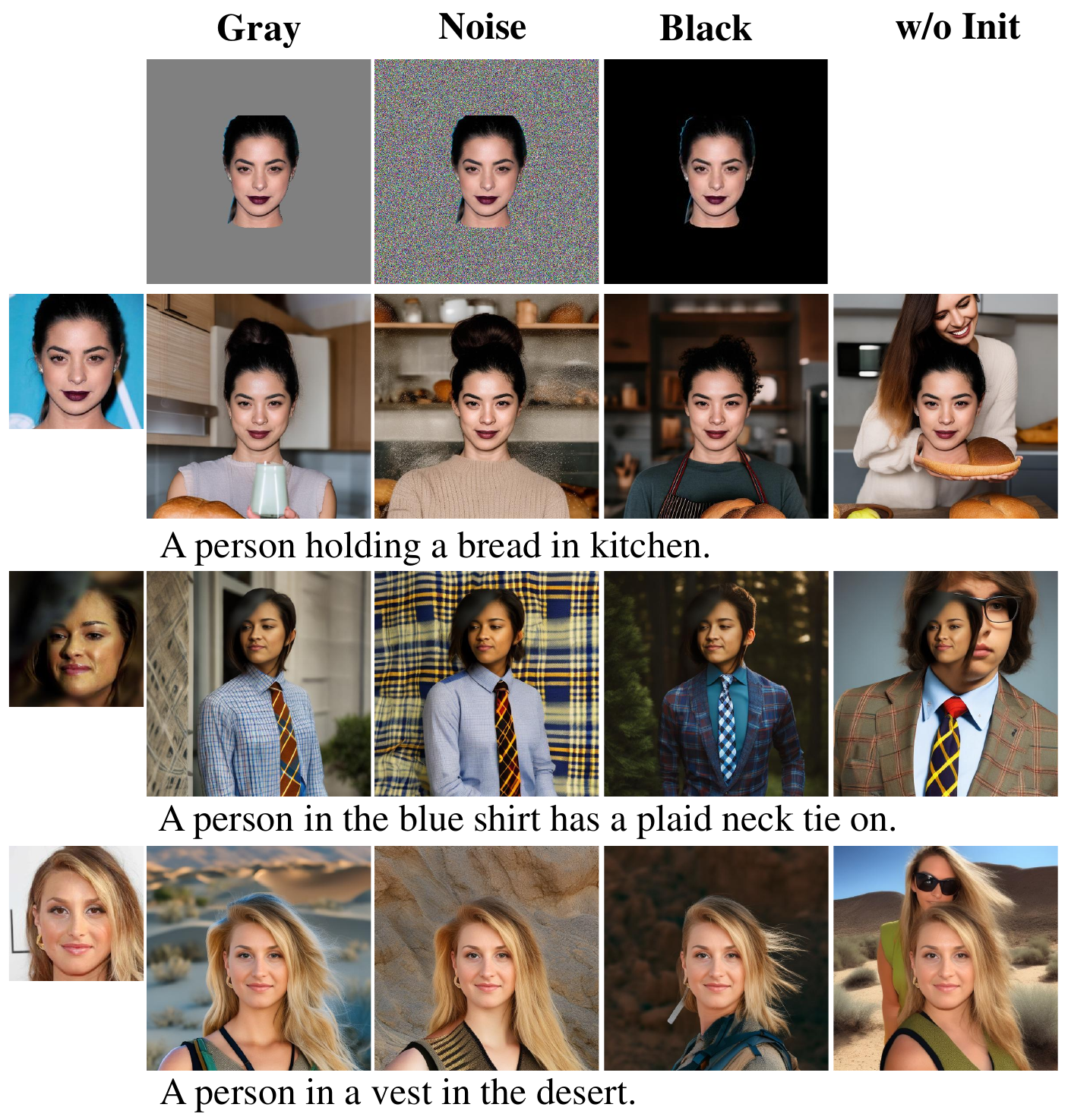}

  \caption{{Ablation study on Seed Initialization.} Given the left-most visual and text conditions, we show generated images with different initial backgrounds (gray, random noise, and black), and without Seed Initialization. }
  \label{fig:seedInitialization}
\end{figure}
\begin{figure}[t]
    \centering
    \includegraphics[width = 0.45\textwidth]{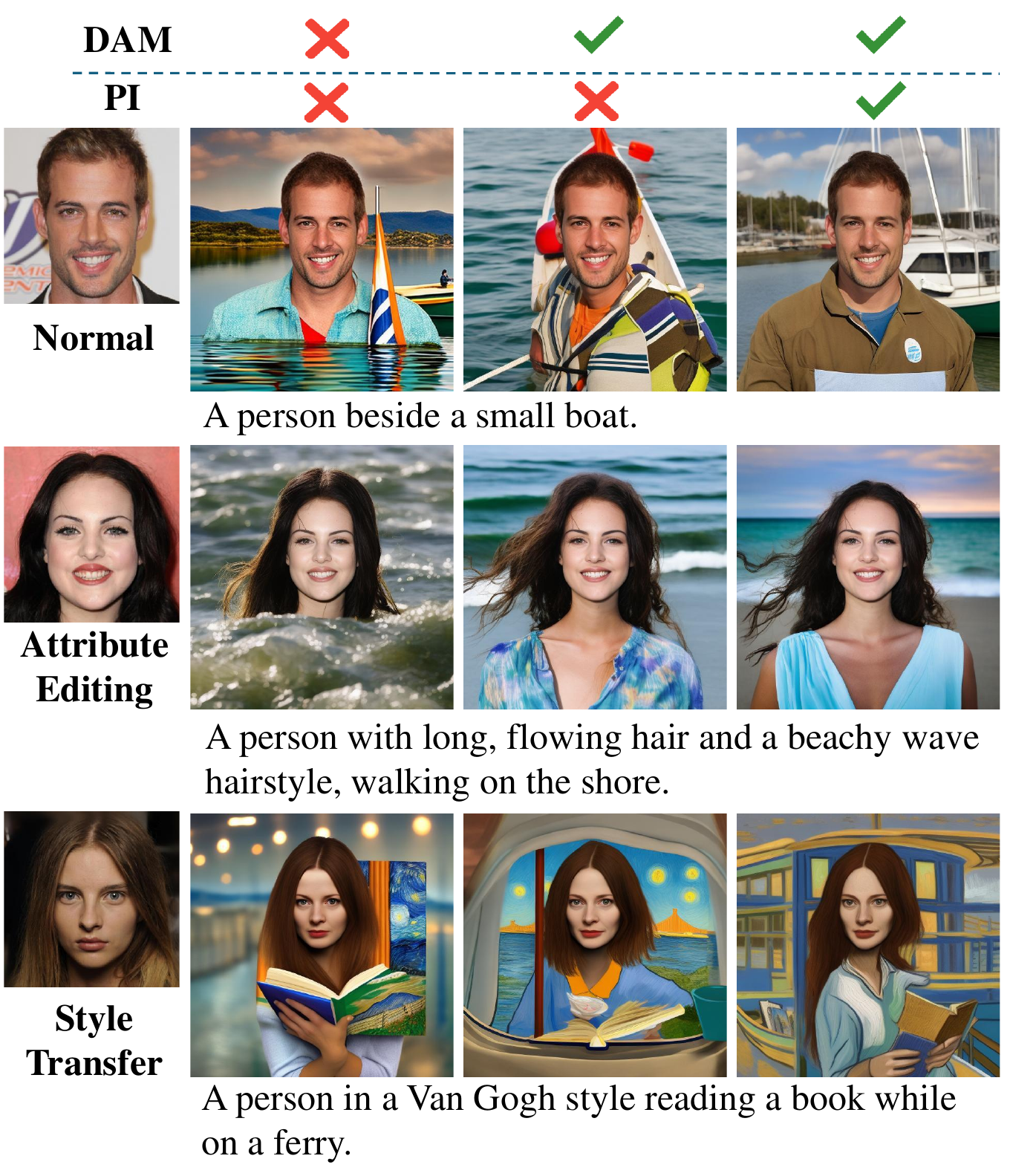}

    \caption{Ablation study of the main improvements in Selective One-Way Diffusion (SOW): dynamic attention modulation (DAM), and prompt intensification (PI). We show the generated image of the three settings.}
    \label{fig:Ablation_comparison}
\end{figure}

\textbf{Quantitative Results.}
Quantitative results in Tab.~\ref{Tab:metric-table} show that our method demonstrated a significant retention rate of 100\% for face detection, indicating its proficiency in preserving the visual condition. In contrast, TI and DreamBooth exhibit significant challenges in preserving facial identities, with face detection rates of 44.50\% and 38.67\%, respectively. While ControlNet is highly time-efficient, it does not effectively preserve identity, evidenced by a poor face detection rate of 96.16\% and a high ID-Distance of 1.092.
However, it is important to acknowledge that SOW does not outperform all methods across every metric. For instance, our SOW scores 0.771 on the ID-Distance metric, which is only worse than SD Inpainting's score of 0.408. SD Inpainting excels in ID preservation due to its enforced region retention; however, this often results in image artifacts and poor human-perceived quality as evidenced by its low condition consistency (7.21\%) and general fidelity (1.25\%), also highlighted in Figure~\ref{fig:teaser}.

\textbf{Human Evaluations.}
To supplement the automatic evaluations, we conducted a preference test in which participants were asked to select the image most consistent with the evaluation criteria\textemdash Condition Consistency and General Fidelity\textemdash from a shuffled set that includes our method and four comparison methods, as illustrated in Table~\ref{Tab:metric-table}. This involved 50 different participants evaluating each of the 600 image groups, resulting in a total of 1,200 responses.
In contrast to DreamBooth, which records the second-highest performance with a Condition Consistency of 25.04\% and a General Fidelity of 32.26\%, SOW demonstrates a significant performance leap, achieving a Condition Consistency of 55.00\%  and a General Fidelity of 52.83\%. This substantial improvement indicates that SOW provides a balanced approach that emphasizes overall image quality and condition consistency, which is crucial for creating visually appealing images.
When processing semantically complex visual conditions, such as faces, by explicitly considering and incorporating low-level visual details, our approach can motivate the model to generate results that are highly consistent with the original conditions. Our method consistently outperforms the others across all three settings, as detailed in Appendix~\ref{sec:image_comparison_with_baselines}.
In addition, our method is \textit{training-free}, thus we can generate the prediction using fewer computations. 
These results demonstrate that our method better integrates text and visual conditions while preserving a satisfying image fidelity.

\subsection{Ablation Studies}
\begin{figure}[!t]
  \centering
  \includegraphics[width=0.5\textwidth]{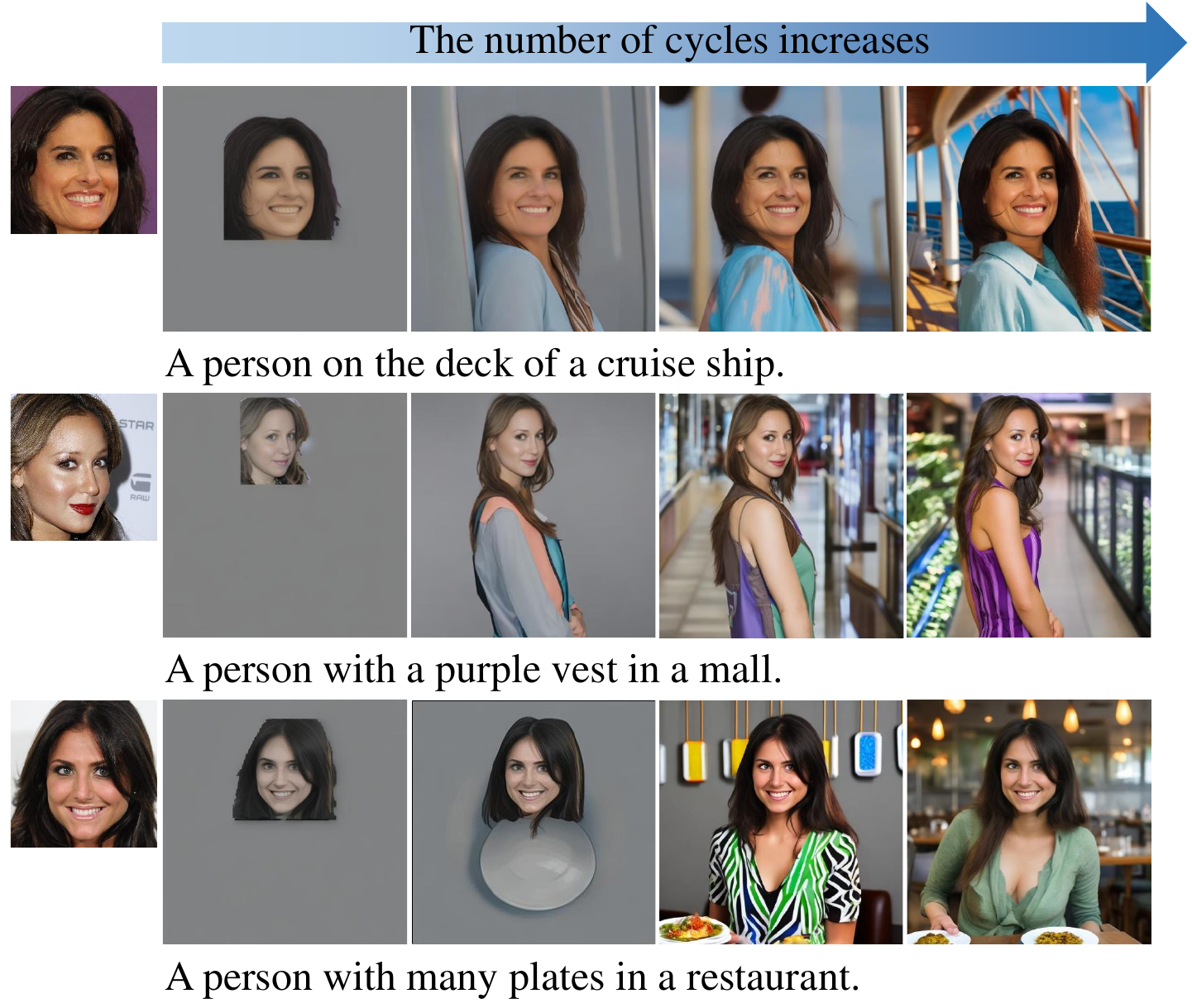}
  \caption{More analysis of the cycling process that diffuses ``visual seed'' to its surroundings. The leftmost figure shows a given face condition. The right shows the images generated with given text conditions. The cycle number increases from the left to the right.}
  \label{fig:backnum_result}
\end{figure}

 \begin{figure}[!t]
    \centering
   
    \includegraphics[width=0.5\textwidth]{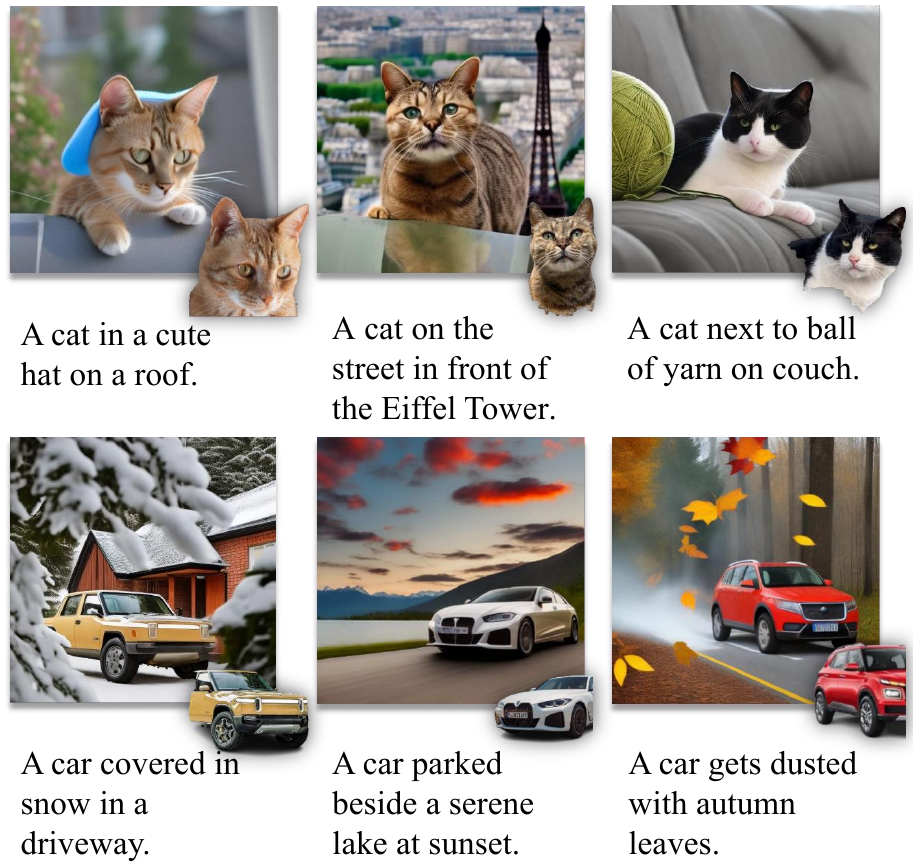}
 
    \caption{For further analysis of generalized applications, we display the generated images conditioned on text alongside the visual input of either part of a cat or a car, which are shown in the bottom right corner of each image.  
    }
 
    \label{fig:generaliazed_cond}
\end{figure}
   
\begin{figure*}[!t]
    \centering
    \includegraphics[width=1\textwidth]{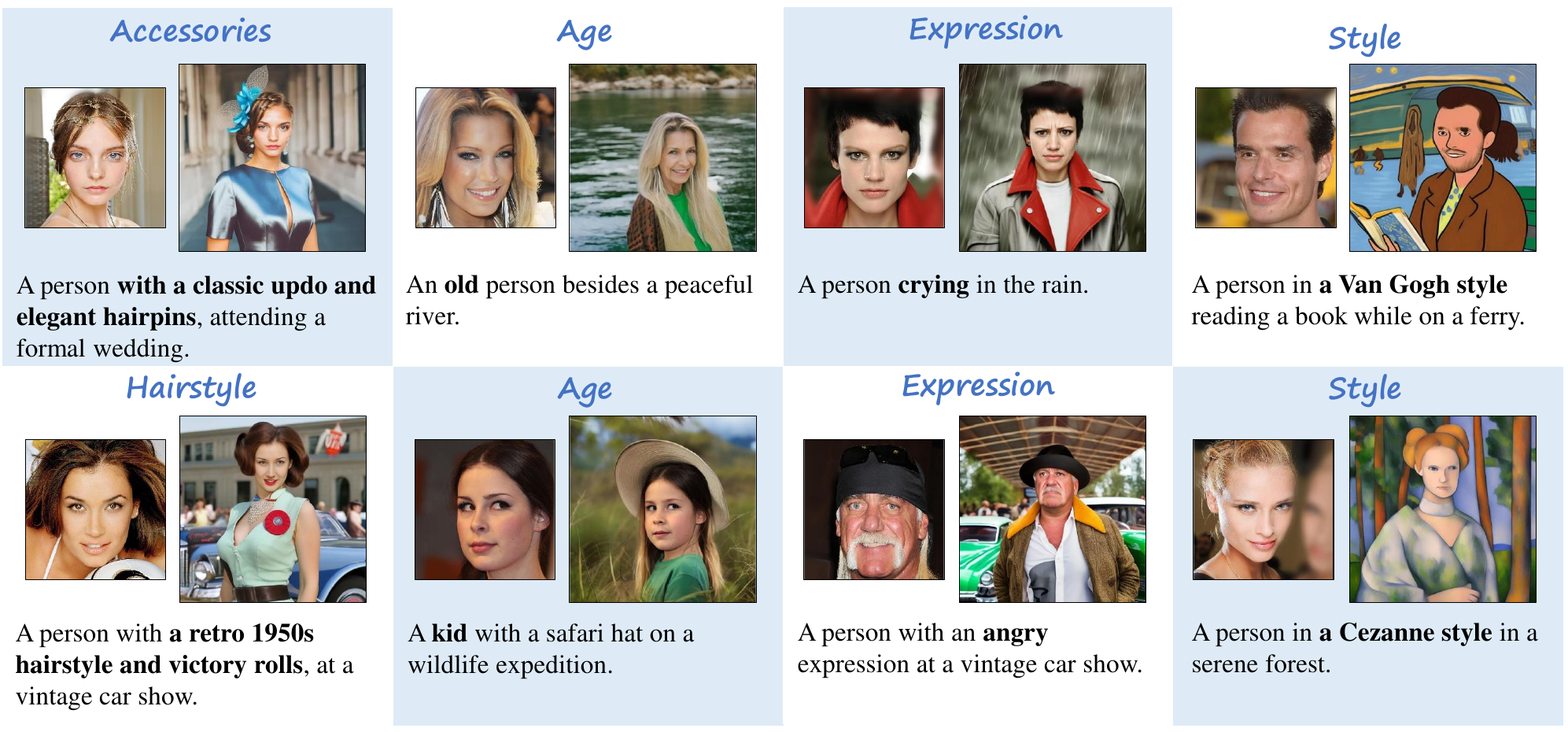}
    \caption{The adaptation of the visual condition to align with the text condition while maintaining the semantic and pixel-level information of the visual condition. In each pair of images, the smaller image is the given visual condition, and the other is the generated result. The bolded parts of the text conditions highlight the conflicts between conditions.}
    \label{adapt_conflict}
\end{figure*}

\textbf{Ablation of Seed Initialization.}
As shown in Fig.~\ref{fig:seedInitialization}, the results without Seed Initialization (SI) exhibit an unnatural merging between the background and visual condition, as another person emerges separately from the visual condition. Without SI, we can not pre-generate the image layout during the Chaos Phase, where the model only fits the textual condition without considering the visual condition. For example, the model may already generate a person at other positions rather than the pre-defined location, which is nearly impossible for us to meet the strict visual condition restrictions given such a layout.
We also try random noise, and black backgrounds for different seed initialization backgrounds for SI. The results show using these backgrounds could slightly decrease the quality of generated images compared to the ones in the gray (default) background. We argue this originates from the fact that the average color value (gray) is a better starting point for all possible synthesized images.

\textbf{Ablation of the Main Improvements in Selective One-Way Diffusion (SOW).}
We conduct an ablation study on main modules\textemdash adaptive position (AP), dynamic attention modulation (DAM), and prompt intensification (PI)\textemdash to validate the effectiveness of the SOW framework. The study comprises the following configurations: 
(a) Using only random positioning (with the same visual condition size restriction as AP), without AP, DAM, or PI; (b) Incorporating AP only; (c) Incorporating both AP and DAM simultaneously; (d) Incorporating AP, DAM, and PI concurrently.

In the quantitative analysis outlined in Tab.~\ref{tab:ablation_o}, we apply the previously established evaluation metrics and introduce an analysis of the occurrence rate of the truncated outlines — where visual conditions appear significantly disconnected from the background, referred to as the failure rate. 
Incorporating the AP (Adaptive Positioning) module allows the MLLM provides reasoned planning based on visual conditions. As a result, compared to configurations without AP, all metrics demonstrate improved performance, leading to better image generation outcomes. The implementation of the dynamic attention modulation leads to a reduction in the truncated outlines by approximately 5.00\% compared to the (b) results, underscoring its effectiveness in enhancing the semantic consistency between the generated images and the accompanying text. Moreover, with the increased application of the prompt intensification module, the ID-Distance decreases to 0.8068 and the incidence of truncated outlines further decreases by about 3.33\% compared to the (c) results, thereby affirming improved semantic consistency between the visual conditions and the generated images.

For qualitative analysis, Fig.~\ref{fig:Ablation_comparison} demonstrates that the integration of the attention modulation module significantly enhances the consistency between the visual condition portion and the overall background. This addition effectively addresses issues of disconnection, particularly the generation of bodies that are coherently connected to heads, thereby substantially mitigating the disconnection problem. The intensified text module further strengthens the model's understanding of the relationship between visual and textual conditions, which in turn improves the alignment between them, leading to the production of higher-quality images.

\textbf{More Ablation of Dynamic Attention Modulation.}
In our refined ablation study, we systematically evaluate the impact of several modifications to dynamic attention modulation on the performance of our model. These modifications included: (a) removing the \(P^+\) operation; (b) removing the \(P^-\) operation; (c) omitting temporal \(\gamma^{time}\) decay; (d) omitting spatial \(\gamma^{dis}\) decay, and (e) eliminating the \(R_{\text{c}}\) judgment. 
First, it is evident that removing either $P^+$ or $P^-$ results in an increase in ID-Distance, and an approximately 6.67\% and 11.67\% increase in the failure rate compared to the original SOW model as shown in Tab.~\ref{tab:ablation_r}. Our designed $P^+$ and $P^-$ operations have a beneficial synergistic effect on overall directional control, contributing to the enhancement of visual condition and overall image consistency.

Further, As illustrated in Tab.~\ref{tab:ablation_r}, the removal of temporal decay (c) results in
a noticeable increase in the average failure rate. This effect is attributed to the constant modulation intensity throughout all iterations, where excessively strong guidance in later stages can disrupt the diffusion model's generative process, adversely affecting the compatibility of visual conditions and overall image quality. Similarly, omitting spatial decay (d) leads to disproportionately strong guidance in areas distant from the visual conditions, resulting in stark contrasts rather than smooth transitions between modulated and unmodulated areas, thereby impairing the model's ability to accurately judge visual conditions and increasing the failure rate. Lastly, the elimination of the indicator $R_{\text{c}}$ (e) allows for continuous dynamic attention modulation at every reverse step. For certain visual conditions that initially align well with the background, applying uniform modulation intensity may alter the original generative trajectory, increasing the ID-Distance and elevating the failure rate.

\subsection{More Analysis}

\textbf{One-Way Diffusion during Cycling Process.}\label{sec:cycle_ablation}
To emphasize the role of cyclic one-way diffusion strategy during generation, we set the noise level $\eta$ to 0 to slow down the information diffusion. As shown in Fig.~\ref{fig:backnum_result}, as the cycle progresses, the background pixels are gradually matched to the text condition and fused with the given face condition. This vividly demonstrates that the information in the visual condition keeps spreading and diffusing to the surrounding region along with cycles. In addition, the model is capable of understanding the semantics of the implanted visual conditions properly.

\textbf{More Applications.}
We directly apply SOW to other applications to show its generalization. 
As shown in Fig.~\ref{fig:generaliazed_cond}, we present results involving a generalized visual condition (part of a cat or a car) paired with different text conditions. Our approach harmoniously extends a complete image from the visual condition under the guidance of the text condition. 

\textbf{Trade-offs between Conditions for Different Modalities.} \label{sec:balance_condition}
In general, in the TV2I task textual information provides a high-level description of the image to be generated, such as the type, color, and other attributes of the object, while visual information contains more low-level details, such as shape, texture, and so on.
The model needs to be able to understand the semantic information in the textual description and translate this information into visually specific details. 
SOW strikes a balance between meeting the visual and text conditions as shown in Fig.~\ref{adapt_conflict}. We showcase a series of samples containing varying degrees of changes in visual conditions of the generated output: add accessories, change the hairstyle, transform the age or the expression, and transfer the style of the image.
For example, when given a photo of a young woman but the text is “an \textit{old} person”, our method can make the woman older to meet the text description by adding wrinkles, changing skin elasticity and hair color, etc. while maintaining the facial expression and the identity of the given woman.
These results demonstrate the ability of SOW to effectively understand and balance the information of different modalities and adaptively adjust to produce high-quality images under a wide range of conditions, showcasing its versatility and effectiveness in handling diverse customization scenarios.

\section{Conclusion and Discussion}\label{sec:conclusion}
In this paper, we delve into the ``diffusion (physics) in diffusion (machine learning)'' properties, proposing an innovative approach termed the \emph{Cyclic One-Way Diffusion} (COW) method. This method transforms the conventional bidirectional diffusion process into a unidirectional one, meticulously directing the inner diffusion. This approach cultivates fertile ground for a wide array of customized application scenarios via a pre-trained yet frozen diffusion model. Building on COW, we further propose a more comprehensive framework \emph{Selective One-Way Diffusion}
(SOW). SOW leverages the formidable reasoning capabilities of Multi-modal Large Language Models (MLLMs) to tackle complex context-driven challenges and utilize dynamic attention modulation in the cyclic process informed by robust prior knowledge of MLLMs,  thereby enhancing the adaptability and efficacy of the generative model. SOW novelly explores the intrinsic diffusion properties tailored to specific task requirements.
% and enhances the adaptability and efficacy of the generative model.  
All the experiments and evaluations demonstrate our method can generate images with high fidelity to both semantic-text and pixel-visual conditions in a training-free, efficient, and effective manner.

\textbf{Limitations.} The pre-trained diffusion model is sometimes not robust enough to handle extremely strong conflicts between the visual and the text conditions. For example, when given the text ``a profile of a person'', and a front face condition, it is very hard to generate a harmonious result that fits both conditions. In this case, the model would follow the guidance of the text generally. 

\textbf{Social Impact.} Image generation and manipulation have been greatly used in art, entertainment, aesthetics, and other common use cases in people's daily lives. However, it can be abused in telling lies for harassment, distortion, and other malicious behavior. Too much abuse of generated images will decrease the credibility of the image. Our work doesn't surpass the capabilities of professional image editors, which mitigates concerns about its potential misuse. Since our model fully builds on the pre-trained T2I model, all the fake detection works distinguishing the authenticity of the image should be able to be directly applied to our results.

\bibliographystyle{IEEEtran}

\bibliography{arxiv}

\clearpage
\newpage

\appendices
% \section{Appendix}
\setcounter{table}{0}
\setcounter{figure}{0}
In the supplementary material, Sec.~\ref{sec:generated_results} showcases an array of TV2I generation results along with comprehensive analyses. 
Furthermore, we conduct extensive qualitative and quantitative image comparisons with baseline methods, and detailed results of human evaluations are elaborated on. 
Our analysis of different hyper-parameters of the number of cycles and positions of the start and end points can be found in Sec.~\ref{appen:hyperparameters}.
Our thorough ablation study shows the effectiveness of dynamic attention modulation and prompt intensification of the three settings (Sec.~\ref{ablation_study}).
Moreover, we illuminate the varying sensitivity to text conditions during the denoising process (Sec.~\ref{sec:text-sensitivity}) and elucidate the semantic formation process of diffusion generation (Sec.~\ref{sec:diff-size_semantic_formation}).

\section{TV2I Generation Results Comparison and Analysis}\label{sec:generated_results}
\subsection{More Comparisons with Baselines}\label{sec:image_comparison_with_baselines}
\begin{table*}[!hb]
  \caption{Quantitative results of objective metrics on the CelebA-TV2I test set and human evaluations under the NORMAL setting.}
  \centering
  \setlength{\aboverulesep}{0pt}
  \setlength{\belowrulesep}{0pt}
  \scalebox{1}{
  \begin{tabular}{l|cc|cc}
    \bottomrule    
    & \multicolumn{2}{c|}{Objective Metrics} & \multicolumn{2}{c}{Human Evaluations} \\ \hline
    Methodology  & ID-Distance $\downarrow$ & Face Detection Rate $\uparrow$ & Condition Consistency $\uparrow$ & General Fidelity $\uparrow$ \\ \hline 
    TI               & 1.182 & 45.00\% & 2.83\% & 9.63\% \\
    DreamBooth       & 1.335 & 40.50\% & \underline{18.77\%} & \underline{21.98\%} \\
    ControlNet       & 1.011 & \underline{99.00\%} & 9.25\% & 2.72\% \\
    SD inpainting    & \textbf{0.389} & \textbf{100.00\%} & 9.25\% & 0.49\% \\
    SOW (ours)       & \underline{0.600} & \textbf{100.00\%} & \textbf{65.19\%} & \textbf{51.87\%} \\
    \bottomrule
  \end{tabular}}
  \label{tab:basic-metrics-normal}
\end{table*}
\begin{table*}[!hb]
  \caption{Quantitative results of objective metrics on the CelebA-TV2I test set and human evaluations under the STYLE TRANSFER setting.}
  \centering
  \setlength{\aboverulesep}{0pt}
  \setlength{\belowrulesep}{0pt}
  \scalebox{1}{
  \begin{tabular}{l|cc|cc}
    \bottomrule    
    & \multicolumn{2}{c|}{Objective Metrics} & \multicolumn{2}{c}{Human Evaluations} \\ \hline 
    Methodology  & ID-Distance $\downarrow$ & Face Detection Rate $\uparrow$ & Condition Consistency $\uparrow$ & General Fidelity $\uparrow$ \\ \hline 
    TI               & 1.212 & 41.50\% & 3.92\% & 13.42\% \\
    DreamBooth       & 1.323 & 41.50\% & \underline{22.79\%} & \underline{35.70\%} \\
    ControlNet       & 1.193 & \underline{92.50\%} & 13.48\% & 5.57\% \\
    SD inpainting    & \textbf{0.400} & \textbf{100.00\%} & 9.07\% & 1.52\% \\
    SOW (ours)       & \underline{1.027} & \textbf{100.00\%} & \textbf{50.74\%} & \textbf{43.80\%} \\
    \bottomrule
  \end{tabular}}
  \label{tab:basic-metrics-style}
\end{table*}

\begin{table*}[!hb]
  \caption{Quantitative results of objective metrics on the CelebA-TV2I test set and human evaluations under the ATTRIBUTE EDITING setting.}
  \centering
  \setlength{\aboverulesep}{0pt}
  \setlength{\belowrulesep}{0pt}
  \scalebox{1}{
  \begin{tabular}{l|cc|cc}
    \bottomrule    
    & \multicolumn{2}{c|}{Objective Metrics} & \multicolumn{2}{c}{Human Evaluations} \\ \hline 
    Methodology  & ID-Distance $\downarrow$ & Face Detection Rate $\uparrow$ & Condition Consistency $\uparrow$ & General Fidelity $\uparrow$ \\ \hline 
    TI               & 1.212 & 42.00\% & 3.55\% & 8.17\% \\
    DreamBooth       & 1.320 & 34.00\% & \underline{33.57\%} & \underline{39.11\%} \\
    ControlNet       & 1.073 & \underline{97.00\%} & 5.20\% & 1.49\% \\
    SD inpainting    & \textbf{0.401} & \textbf{100.00\%} & 3.31\% & 1.73\% \\
    SOW (ours)       & \underline{0.686} & \textbf{100.00\%} & \textbf{54.37\%} & \textbf{49.50\%} \\
    \bottomrule
  \end{tabular}}
  \label{tab:basic-metrics-editing}
\end{table*}

\begin{figure*}[!t]
    \centering
    \includegraphics[width=0.7\textwidth]{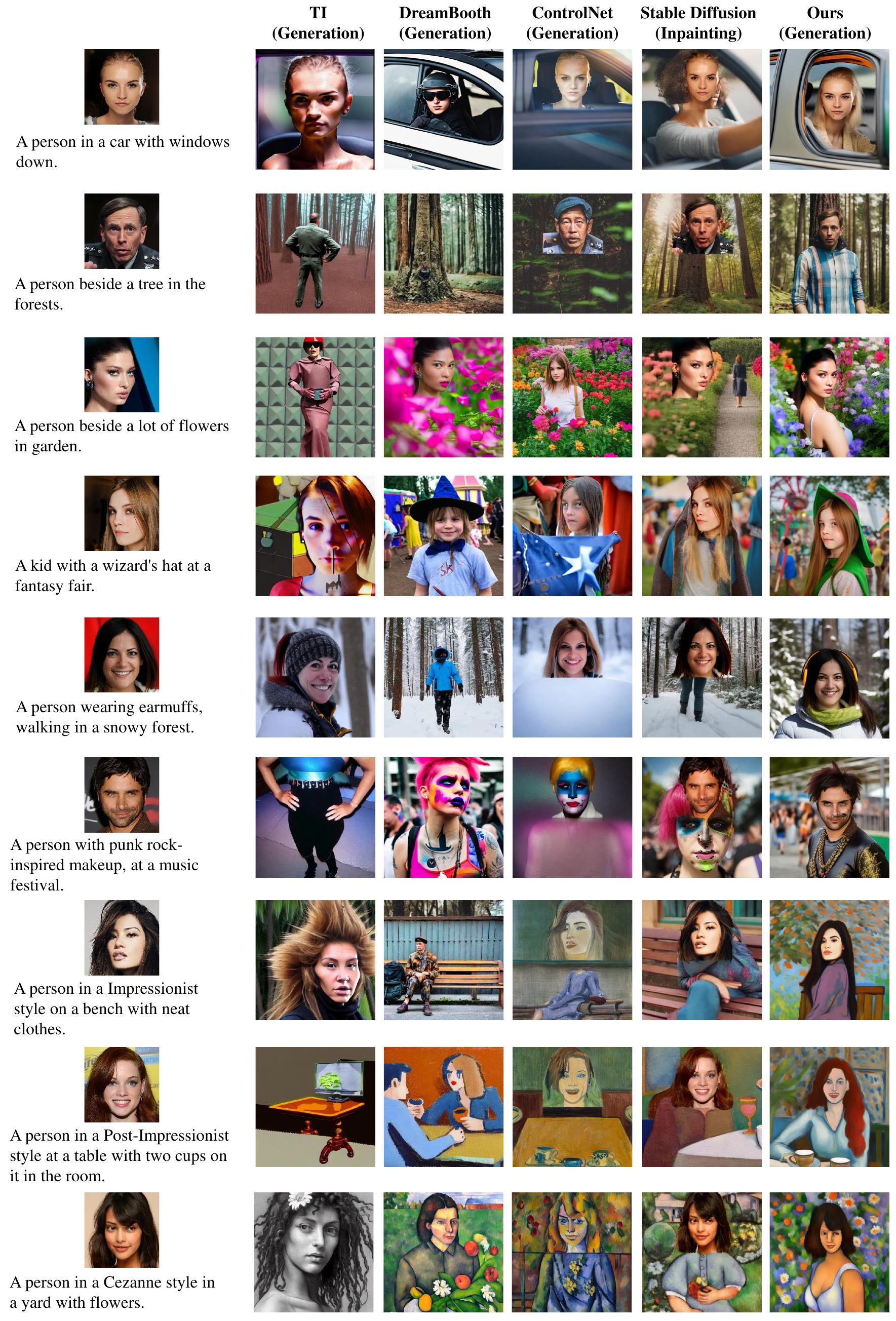}
    \caption{Comparison of our SOW-generated images with TV2I baselines.}
    \label{compare}
\end{figure*}

In Fig.~\ref{compare}, we include a more intuitive comparison with baseline results.
We compare current methods that incorporates visual conditions into pre-trained T2I models: DreamBooth~\cite{ruiz2022dreambooth}, TI~\cite{gal2022ti}, ControlNet~\cite{controlnet}, and SD inpainting~\cite{rombach2022ldm}. As shown in Tab.~\ref{tab:basic-metrics-normal}, Tab.~\ref{tab:basic-metrics-style}, and Tab.~\ref{tab:basic-metrics-editing}, our method works well and outperforms most baselines in most evaluation metrics across the three settings.

We conduct a human evaluation based on two criteria: 1) {Condition Consistency}: whether the generated image well matches both the visual and textual conditions; 2) {General Fidelity}: whether the chosen image looks more like a real image in terms of image richness, face naturalness, and overall image fidelity.  It is worth noting that we inform the participants of the application setting: input the face text condition pairs and then get an output image, and ask them to choose a favorite result if they were the users under the two basic criteria. The voting rate for our results greatly surpasses the other baselines in both criteria across all three settings, as shown in Tab.~\ref{tab:basic-metrics-normal}, Tab.~\ref{tab:basic-metrics-style}, and Tab.~\ref{tab:basic-metrics-editing}, and this demonstrates the general quality when applied to TV2I generation. 

 \begin{table}[!h]
    \centering
    \caption{Comparison of different hyper-parameters of the cycle number and the cycle position on the CelebA-TV2I validation set.}
    \label{Tab:different_hyper-parameters.}
    \vspace{0mm} 
    \setlength{\aboverulesep}{0pt}
    \setlength{\belowrulesep}{0pt} 
    \resizebox{0.5\textwidth}{!}{
    \begin{tabular}{cc|ccc}
    \toprule
    Cycle & Position & ID-Distance $\downarrow$ & Face Detection Rate $\uparrow$ & Time Cost $\downarrow$ \\ \hline \hline
    10 & 700 $\rightarrow$ 500 & 0.8068 & 100.00\% & 4.67 \\
    \hdashline
    10 & 900 $\rightarrow$ 700 & 1.0520 & 96.67\% & 4.93 \\
    10 & 800 $\rightarrow$ 400 & 0.7194 & 98.33\% & 9.59\\
    10 & 500 $\rightarrow$ 300 & 0.6606 & 98.33\% & 5.08\\
    \hdashline
    5 & 700 $\rightarrow$ 500 & 0.6726 & 100.00\% & 2.45 \\
    20 & 700 $\rightarrow$ 500 & 0.6714 & 100.00\% & 9.65 \\
    \bottomrule
    \end{tabular}
    }
\end{table}
\subsection{More Analysis of Different Hyper-parameters}
\label{appen:hyperparameters}
We conduct an analysis of different hyper-parameters of SOW, including the number of cycles, or positions of the start and end points in Tab.~\ref{Tab:different_hyper-parameters.}.
Here we use three quantitative metrics to analyze different models. 

Since different starting and ending points would involve different inversion steps, the time cost is slightly different according to the diffusion inversion cost. 
The results show that the model's performance is sensitive to the starting and ending points, which confirms our motivation to repeatedly go through the semantic formation stage to have a controllable and directional generation.
As we can see, a cycle number of 10 is sufficient and achieves a good balance between performance and speed for the inner diffusion process.

\subsection{More Ablation Study results}
\label{ablation_study}

As shown in Tab.~\ref{tab:overall_ablation_analysis_normal}, Tab.~\ref{tab:overall_ablation_analysis_style}, and Tab.~\ref{tab:overall_ablation_analysis_editing} we conduct ablation study of main improvements on different settings. The modules AP (Adaptive Position), DAM (Dynamic Attention Modulation), and PI (Prompt Intensification) effectively maintain consistency in the generated images across all three settings.

\begin{table}[!htbp]
\setlength{\tabcolsep}{4pt}
\caption{Ablation study of the main improvements under the NORMAL setting: adaptive position (AP), dynamic attention modulation (DAM), and prompt intensification (PI) on the CelebA-TV2I validation set.}
\vspace{5mm} 
\centering
\setlength{\aboverulesep}{0pt} 
\setlength{\belowrulesep}{0pt} 
\scalebox{1}{
\begin{tabular}{ccc|cc}
    \toprule
     AP & DAM & PI & ID-Distance $\downarrow$ & Failure Rate $\downarrow$ \\ \hline \hline
    \xmark & \xmark & \xmark & 0.628 & 20.00\% \\
    \cmark & \xmark & \xmark & 0.643 & 10.00\% \\
    \cmark & \cmark & \xmark & 0.621 & 5.00\% \\
    \cmark & \cmark & \cmark & \textbf{0.604} & \textbf{0.00\%} \\
    \bottomrule
\end{tabular}}
\vspace{-1mm} 
\label{tab:overall_ablation_analysis_normal}
\end{table}

\begin{table}[!htbp]
\setlength{\tabcolsep}{4pt}
\caption{Ablation study of the main improvements under the STYLE TRANSFER setting: adaptive position (AP), dynamic attention modulation (DAM), and prompt intensification (PI) on the CelebA-TV2I validation set.}
\vspace{5mm} % Adjust vertical space above the table
\centering
\setlength{\aboverulesep}{0pt} 
\setlength{\belowrulesep}{0pt} 
\scalebox{1}{
\begin{tabular}{ccc|cc}
    \toprule
      AP & DAM & PI & ID-Distance $\downarrow$ & Failure Rate $\downarrow$ \\ \hline \hline
    \xmark & \xmark & \xmark & 0.822 & 10.00\% \\
    \cmark & \xmark & \xmark & \textbf{1.098} & 20.00\% \\
    \cmark & \cmark & \xmark & 1.122 & \textbf{10.00\%} \\
    \cmark & \cmark & \cmark & \textbf{1.098} & \textbf{10.00\%}\\
    \bottomrule
\end{tabular}}
\vspace{-1mm} 
\label{tab:overall_ablation_analysis_style}
\end{table}

\begin{table}[!htbp]
\setlength{\tabcolsep}{5pt}
\caption{Ablation study of the main improvements under the ATTRIBUTE EDITING setting: adaptive position (AP), dynamic attention modulation (DAM), and prompt intensification (PI) on the CelebA-TV2I validation set.}
\vspace{5mm} % Adjust vertical space above the table
\centering
\setlength{\aboverulesep}{0pt}
\setlength{\belowrulesep}{0pt}
\scalebox{1}{
\begin{tabular}{ccc|cc}
    \toprule
      AP & DAM & PI & ID-Distance $\downarrow$ & Failure Rate $\downarrow$ \\ \hline \hline
    \xmark & \xmark & \xmark & 0.881 & 20.00\% \\
    \cmark & \xmark & \xmark & 0.730 & 10.00\% \\
    \cmark & \cmark & \xmark & \textbf{0.684} & 10.00\% \\
    \cmark & \cmark & \cmark & 0.718 & \textbf{5.00\%} \\
    \bottomrule
\end{tabular}}
\vspace{-1mm} 
\label{tab:overall_ablation_analysis_editing}
\end{table}

\section{Properties of Diffusion Generation Process} \label{sec:properties}
\begin{figure}[!h]
    \centering
    \includegraphics[width=0.5\textwidth]{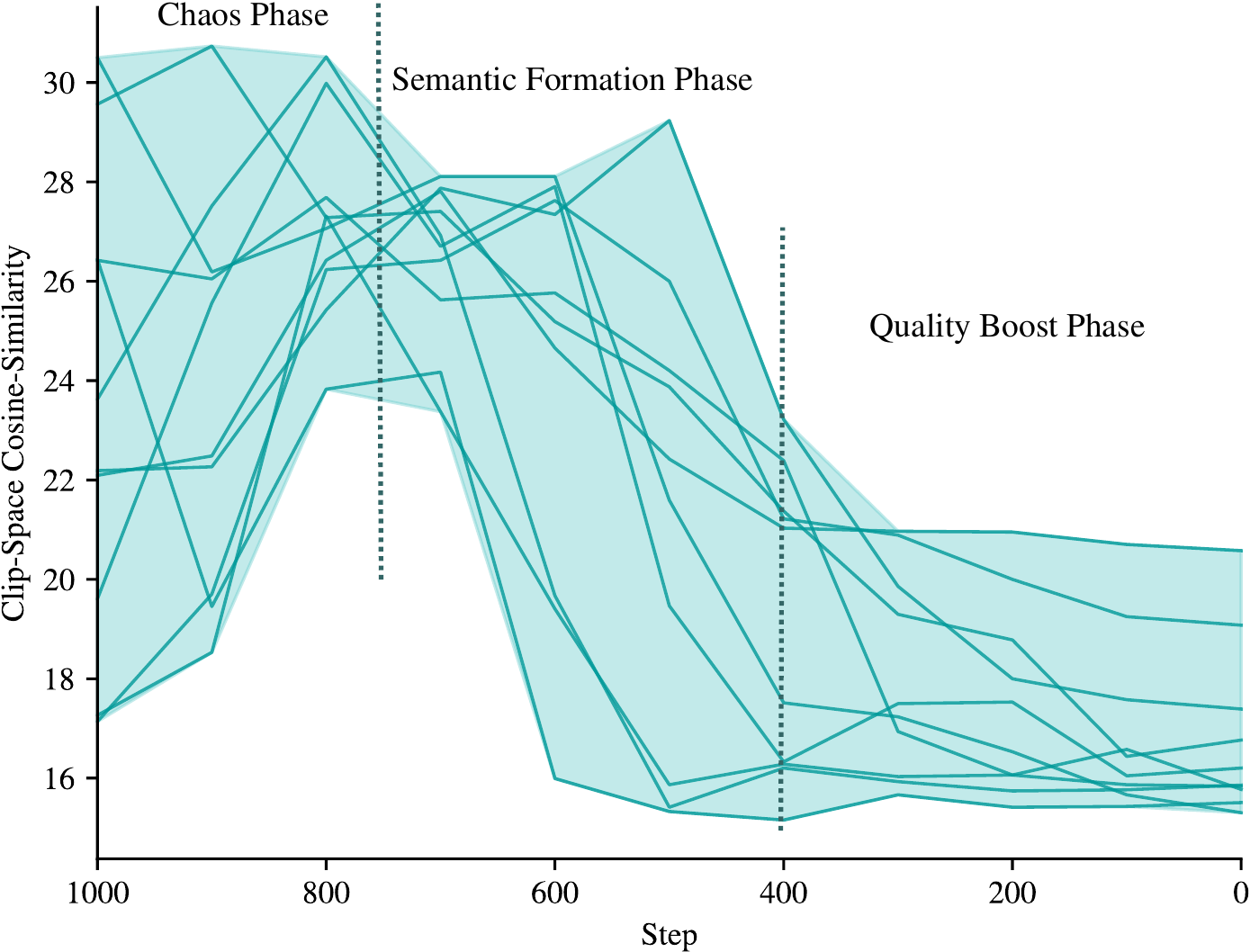}
    \caption{Text-sensitivity during denoising process. Each line represents a CLIP cosine similarity between the generated image and text with the text condition injected at different steps. The image is generated in 1000 overall unconditional denoising process with 100 steps text conditional guidance starting from $t$. Generally, the denoising process is more responsive to the text condition in the beginning and almost stops reacting to the text condition when high-level semantics are settled.}
    \label{text_sensitivity}
\end{figure}
\subsection{Sensitivity to Text Condition during Denoising}\label{sec:text-sensitivity}
We study the sensitivity of diffusion models to text conditions during the denoising process by injecting text conditions at different denoising steps. We replace a few unconditional denoising steps with text-conditioned ones during the generation of the image randomly sampled from Gaussian noise. We calculate the CLIP cosine similarity between the final generated image and text condition. As shown in Fig.~\ref{text_sensitivity}, the model is more sensitive to text conditions in the early denoising stage and less sensitive in the late stage. This also demonstrates our method reasonably utilizes the sensitivity to text conditions through proposed Seed Initialization and Cyclic One-Way Diffusion.

\subsection{Illustration of Semantic Formation Process of Diffusion Generation}\label{sec:diff-size_semantic_formation}

\begin{figure}[!h]
    \centering
    \includegraphics[width=0.45\textwidth]{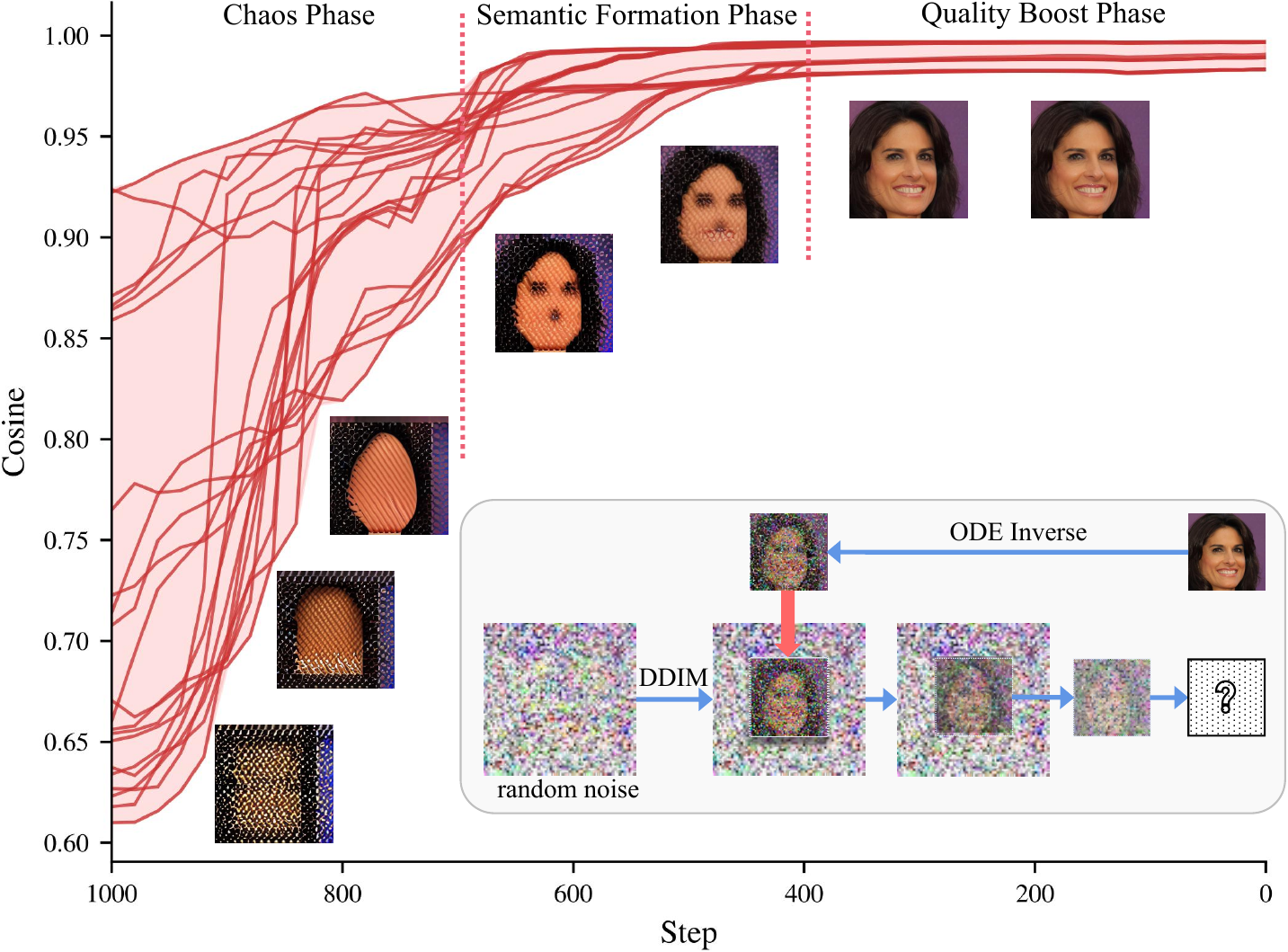}
    % \vspace{-3mm}
    \caption{Illustration of the semantic formation process and the mutual interference. Each red curve represents a face disturb-and-reconstruct process. We only show one face for instance. We disturb the reconstruction process by sticking the origin image to a random noise background at different steps. The cosine similarity between the original image and the reconstructed image increases as the denoising step goes. 
    }
    \label{fig:measure_influence_along_denoise_process}
\end{figure}

To quantitatively assess the influence between regions during the denoising process, we inverse an image $x_0$ to obtain different transformed $\mathbf{x_t}$, and plant it in a replacement manner into a sub-region of random Gaussian noise in the same step. We subsequently apply this composite noise map to several denoising steps before extracting the planted image to continue the generation process. As shown in Fig.~\ref{fig:measure_influence_along_denoise_process}, the influence level weakens with the denoising process. 

\begin{figure}[!h]
    \centering
    \includegraphics[width=0.5\textwidth]{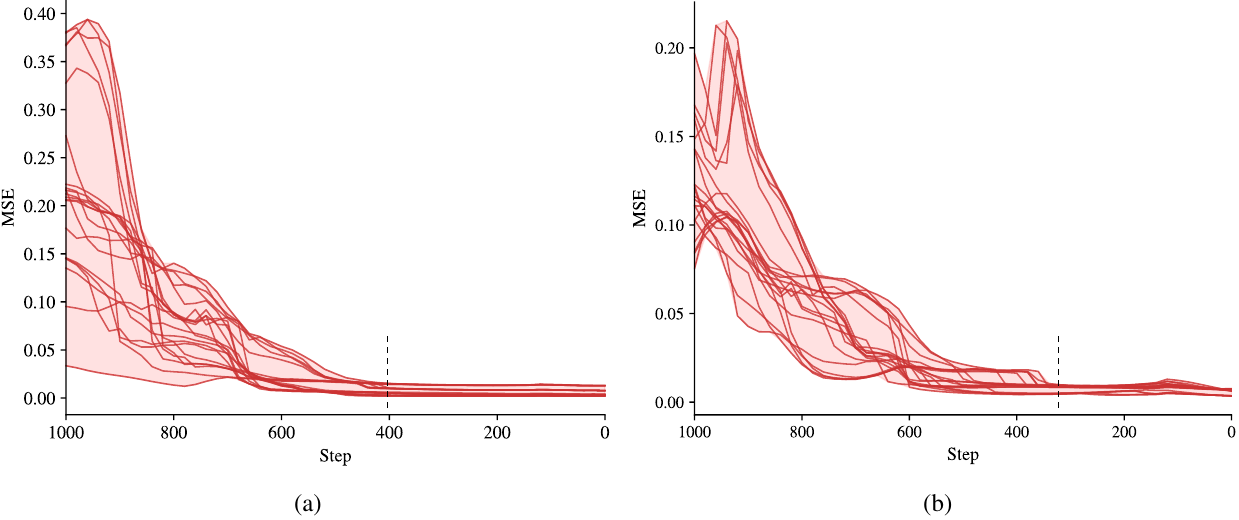}
    \caption{Different sizes come with different semantic formation processes. Each red curve represents a face disturb-and-reconstruct process, (a) size is $256 \times 256$ and (b) size is $128 \times 128$. We disturb the reconstruction process by sticking the origin image to a random noise background ($512 \times 512$) at different steps. The general semantic is settled earlier when the size is larger.}
    \label{fig:size-mse}
\end{figure}

To see the degree of impact exerted on images of different sizes during the denoising process, we conduct the same experiments with images of two different sizes ($128\times128$, $256\times256$). We inverse images to $\mathbf{x_t}$, and reconstruct them with disturbance introduced by sticking them to a random noise background ($512 \times 512$) at the corresponding step for 100 steps. We calculate the MSE loss of the original image and the final reconstructed image after being disturbed. As shown in Fig.~\ref{fig:size-mse}, the semantics settle earlier when the size is larger.

\vspace{-1cm}
\vspace{-10cm}

\vfill

\end{document}